\documentclass[runningheads]{llncs}

\usepackage{eccv}

\usepackage{eccvabbrv}
\usepackage{hyperref}
\usepackage{booktabs}
\usepackage{longtable}
\usepackage{multirow}
\usepackage{graphicx}
\usepackage{makecell}  
\usepackage[table]{xcolor}
\usepackage{fancybox} 
\usepackage{wrapfig}

\newcommand{\Finding}[3]{%
  \noindent\ovalbox{%
    \parbox{0.96\linewidth}{%
      \textbf{Takeaway #1: }%
      \colorbox{green!25}{\strut #2}:\ #3%
    }%
  }%
  \vspace{0.8em}%
}

\usepackage[accsupp]{axessibility}  

\renewcommand{\arraystretch}{1.05}
\usepackage{orcidlink}

\begin{document}

\title{On Test-Time Scaling for Vision-Language Models} 


\author{Fawaz Sammani\inst{1,2} \and
Tzoulio Chamiti\inst{1,2} \and
Nikos Deligiannis\inst{1,2}}

\authorrunning{F. Sammani et al.}

\institute{
ETRO Department, Vrije Universiteit Brussel, Belgium \and
imec, Kapeldreef 75, B-3001 Leuven, Belgium\\
\email{\{fawaz.sammani, tzoulio.chamiti, nikos.deligiannis\}@vub.be}
}

\maketitle

\begin{abstract}
Test-time scaling is a paradigm where large models use additional compute at inference to achieve better performance, without changing model weights. While it has been widely studied for Large Language Models (LLMs), its applicability to Large Vision-Language Models (LVLMs) remains less explored and analyzed, with limited analysis of whether, when, and to what extent these approaches transfer to LVLMs. In this work, we ask a simple but fundamental question: can conventional test-time scaling methods developed for LLMs be directly applied to LVLMs? We present the first comprehensive study of test-time scaling for LVLMs, spanning multiple models and model sizes, nine test-time scaling methods, and six diverse benchmarks. Our main findings is that 1) different from previous findings, small, well-performing models benefit the most from test-time scaling, enabling performance improvements of up to around 30\%, reaching large models performance, and often outperforming them, 2) LVLMs lose focus when given more compute than necessary, and 3) Visual information is encoded early in the reasoning chain, after which the chain is dominated by text-only reasoning and the contribution of image tokens drops significantly. Finally, we also provide a global and fine-grained analysis on the quality and information sufficiency of the reasoning chains produced. Overall, our findings and analysis provide practical guidance and insights into LVLMs and their deployment in research and industry.
\end{abstract}

\begin{figure}
    \centering
    \includegraphics[width=1\linewidth]{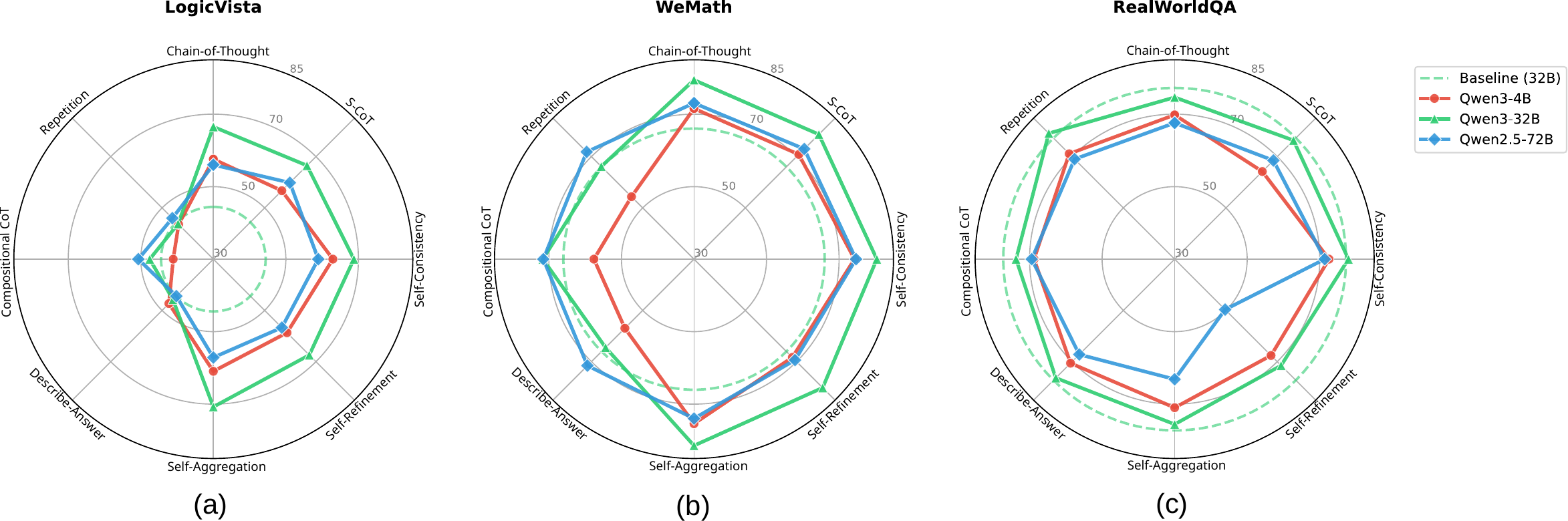}
    \caption{Radar plot showing 8 test-time scaling methods on three benchmarks: reasoning (LogicVista, WeMath) and perception (RealWorldQA) for Qwen3-VL-4B, Qwen3-VL-32B and Qwen2.5-VL-72B. The dashed circle indicates baseline performance without test-time scaling, while solid lines correspond to test-time scaling strategies.}
    \label{fig:teaser_fig}
\end{figure}

\section{Introduction}
\label{sec:intro}
Test-time scaling (TTS) has emerged as a powerful way to improve model performance by allocating more computation at inference time, while keeping the model parameters fixed. It offers an alternative to parameter scaling (i.e., making models larger), delivering substantial accuracy gains on reasoning tasks, and recent evidence suggests that, for Large Language Models, test-time scaling can be more effective than increasing parameter count \cite{snell2025scaling}. 

\noindent At its core, test-time scaling expands the model search space by eliciting and activating reasoning capabilities already present in the model \cite{yue2025does}. Current approaches mainly fall into two paradigms. The first relies on additional training: Large Language Models (LLMs) or Large Vision–Language Models (LVLMs) are fine-tuned on reasoning data via supervised fine-tuning, often followed by reinforcement learning \cite{Muennighoff2025s1ST, Ou2025EmpoweringLM, Suma2025DeepSeekR1IR, Xu_2025_ICCV, zhang-etal-2025-improve, Zhang2023MultimodalCR, chen2025sft, yao2025rsharevl}. Models trained in this way are commonly referred to as "thinking models". The second paradigm achieves test-time scaling without further training, instead leveraging the instruction-tuned model and activating its reasoning behavior through prompting \cite{wei2022chain, plansolve}, sampling \cite{selfconsistency}, aggregation \cite{selaggregate}, or iterative refinement \cite{selfrefine}. These methods are often described as \textit{zero-shot} because they activate reasoning behavior without further training. Motivated by recent evidence that strong reasoning capabilities can already reside in instruction-tuned models \cite{yue2025does, Venkatraman2025RecursiveSU}, our work focuses on this second paradigm.
\newline
\noindent While test-time scaling has been extensively studied for LLMs, it remains less explored and analyzed for LVLMs, especially across different model scales and benchmarks. As a result, it is still unclear whether, when, and to what extent these approaches transfer to LVLMs. Prior work on LLMs \cite{Lanham2023MeasuringFI, Ou2025EmpoweringLM, magister-etal-2023-teaching} report that test-time scaling is  ineffective for small LLMs (<10B): it yields no accuracy gain, and prompting them to reason can even reduce accuracy. Recent work \cite{kaya2026efficient} evaluates several test-time scaling methods on small vision-language models and reports a similar trend. Other studies also suggest that Chain-of-Thought prompting (i.e., prompting the model to "think step-by-step") fails to improve performance in LVLMs \cite{ccot}. This led researchers to develop test-time scaling techniques designed specifically for VLMs, under the assumption that test-time scaling methods for LLMs do not transfer effectively to VLMs.

\noindent In this work, we revisit these assumptions and ask a simple but fundamental question: can conventional test-time scaling methods developed for LLMs be directly applied to LVLMs? Our results indicate that the answer is yes—and that these methods are highly effective. We show that standard test-time scaling strategies from the LLM literature transfer well to LVLMs, and especially to small yet capable vision-language models. Interestingly, we find that small models often benefit substantially more from test-time scaling than larger ones, with performance improvements of up to 30\%, reaching close-to baseline performance of large models, and in some cases, even outperforming them. Figure \ref{fig:teaser_fig} presents a radar plot comparing three different benchmarks across nine methods and three models. Figures \ref{fig:teaser_fig}(a,b) show that the smaller model Qwen3-VL-4B, when combined with test-time scaling, reaches and even surpasses the performance of the larger Qwen3-VL-32B baseline without test-time scaling (indicated by the dashed circle). We also find that these conventional test-time scaling methods significantly perform better than approaches developed specifically for vision-language models (e.g., Compositional CoT \cite{ccot}). 

\noindent On the other hand, we find that test-time scaling can hurt performance on perceptual tasks (i.e., tasks that do not require reasoning), rather than maintaining their performance (Figure \ref{fig:teaser_fig}c). By analyzing token budgets, we observe that VLMs "loose focus" when given more compute than necessary; they tend to overthink, make wrong assumptions, and accumulate errors that ultimately lead to hallucinations and incorrect final answers. We further support this conclusion with an in-depth attention analysis of LVLMs. Specifically, we find that overly verbose generations rely heavily on early visual encoding; there is a short initial window during which generated tokens attend to the image and encode visual information into the hidden state. After this window, the model increasingly relies on that hidden representation, while attention to the image gradually decays over the remainder of the generation.
\newline
\noindent As a final contribution, we analyze the quality and information sufficiency of the generated reasoning chains at both global and fine-grained levels, using an LLM-as-a-judge approach and human evaluation. 

\section{Related Work}
Early approaches to reasoning in vision and vision-language tasks \cite{Sammani2022NLXGPTAM, Sammani2023UniNLXUT} trained models to generate both an answer and a single reasoning phrase within a unified output text stream. This was later extended to zero-shot settings \cite{sammani2025zeroshot}. With the rise of large language models (LLMs), several techniques have been introduced to improve their performance via prompting \cite{wei2022chain, plansolve}, sampling and aggregation \cite{selfconsistency, selaggregate}, refining the model's own answers through iterative feedback \cite{selfrefine} and others. The core idea of these techniques is to spend more compute at inference time before producing the final answer. This is commonly referred to as test-time scaling. Several works also introduce approaches for test-time scaling tailored specifically for vision–language models. For example, CCoT \cite{ccot} extracts a scene graph from the input image and injects it back into the prompt. DCoT \cite{jia2024dcot} performs multi-pass prompting by first extracting bounding-box-based visual cues and then conditioning subsequent steps on these localized signals. DDCoT \cite{zheng2023ddcot} separates recognition and reasoning into sequential prompting stages, while \cite{kaya2026efficient} aggregates token output distributions over augmented image-text pairs. 
\newline
\newline
Most of these works report that Chain-of-Thought prompting and other inference-time scaling methods fail and even yield negative gains for LVLMs \cite{kaya2026efficient, ccot}, mirroring findings for small LLMs \cite{Lanham2023MeasuringFI, magister-etal-2023-teaching, Ou2025EmpoweringLM}. In contrast, we revisit this conclusion by systematically testing whether, when and to what extent do classic LLM test-time scaling methods transfer directly to LVLMs, and we examine this question across model scales and generations and across different benchmarks, coupled with deep analysis of attention and reasoning-chain sufficiency.

\section{Test-Time Scaling of Vision-Language Models}
\label{section:method}

A LVLM typically consists of (1) A vision encoder that encodes an image $I \in \mathbb{R}^{3\times H \times W}$ into a sequence of visual tokens $P_v \in \mathbb{R}^{N \times d}$, where $N$ is the number of visual tokens and $d$ is the dimensionality of each token, (2) a pretrained Large Language Model (LLM), and (3) an adapter that projects the visual tokens $P_v$ into the LLM’s language embedding space. Some LVLMs additionally compress the visual tokens to reduce the token count. We denote the adapter output by $P_v^{ad} \in \mathbb{R}^{N_m \times d}$, where $N_m$ is the number of (compressed) visual tokens that enter to the LLM. For example, in the Qwen series, $N_m = N / 4$. The task of the LVLM is to respond to an instruction provided by the user. For Visual Question Answering (VQA), the instruction is a \texttt{question} about the image. The \texttt{question} can be (i) an open-ended question, where the model must answer with a single word or phrase; (ii) a multiple-choice question (MCQ), where the model must select one option from a provided list; or (iii) a passage followed by a question. We include a \texttt{post-prompt} to instruct the LVLM to answer in a structured format. The full text prompt is therefore $\{\texttt{question}\};\{\texttt{post-prompt}\}$, where ";" indicates string concatenation (see Section \ref{supp:post_prompts} of the Supp. material for exact details on \texttt{post-prompt}). This prompt is tokenized into a sequence of text tokens $P_t \in \mathbb{R}^{T \times d}$, where $T$ is the total number of text tokens and $d$ is the token dimensionality. The total input sequence $P$ to the LVLM is the concatenation of the visual and textual tokens, i.e., $P \in \mathbb{R}^{(N_m+T) \times d}$. The LVLM outputs the task \texttt{answer}, which may be a single letter (for multiple-choice questions) or a single word or phrase (for open-ended questions). 
\newline
\newline
\noindent \textbf{Methods:} We consider a large suite of test-time scaling methods, some borrowed from the LLM literature and others specifically designed for LVLMs. All these methods are applied at inference time to elicit and activate reasoning capabilities of the LVLM, without changing its weights or performing any additional training. All techniques below are implemented via appending a \texttt{think-prompt} before \texttt{post-prompt}. For example, for Chain-of-Thought prompting, the \texttt{think-prompt} is "think step-by-step". The final input prompt $P_t$ is therefore
\begin{center}
\{\texttt{question}\};\{\texttt{think-prompt}\};\{\texttt{post-prompt}\}.
\end{center}
The LVLM outputs a reasoning chain $r$ followed by the \texttt{answer}. Note that although the \texttt{post-prompt} instructs the LVLM to output \texttt{answer} in a specific format, the model can still occasionally fail to comply. To reliably extract the answer in such cases, if the answer extraction fails, we perform a second run by providing
\begin{center}
\{\texttt{question}\};\{\texttt{r}\};\{\texttt{answer-only}\};\{\texttt{post-prompt}\}.
\end{center}
where \texttt{answer-only} prompts the LVLM to answer \texttt{question} given $r$. We consider the techniques below, and provide the \texttt{think-prompt} for each method in Section \ref{supp:thinking_prompts} of the Supp. material.
\newline
\noindent \textbf{\textcircled{1} Chain-of-Thought (CoT)} prompting \cite{wei2022chain} elicits the LVLM to generate intermediate reasoning steps before producing a final answer by prompting it to think step by step. This method uses a single forward pass, but consumes more “thinking” tokens.

\noindent \textbf{\textcircled{2} Structured Chain-of-Thought (S-CoT)} prompting is a more structured variant of \textcircled{1} used to guide the LVLM through question decomposition, information organization, logical reasoning, and final summarization.

\noindent \textbf{\textcircled{3} Plan-and-Solve (PaS)} prompting \cite{plansolve} first asks the model to understand and devise a plan for the problem and then to execute that plan step by step to produce the final answer. Like CoT and S-CoT, this method uses a single forward pass, but consumes more “thinking” tokens.

\noindent \textbf{\textcircled{4} Self-Consistency} \cite{selfconsistency} samples multiple independent CoTs (i.e., applies \textcircled{1} for $b$ runs), producing $b$ different reasoning chains for the same problem, and then selects the most frequent answer as the final response (see Section \ref{supp:method_params} of the Supp. material for sampling hyperparameters). This technique promotes diversity in intermediate reasoning, helping the LVLM explore richer solution paths. It uses $b$ forward passes, each producing more “thinking” tokens.

\noindent \textbf{\textcircled{5} Self-Aggregation} \cite{selaggregate} is similar to \textcircled{4} with one key difference; it first produces $b$ different reasoning chains, but instead of discarding each and keeping only its final answer, it retains all $b$ sampled chains and their answers, concatenates them into a single chain, and prompts the LVLM to aggregate the diverse information across chains to produce a final reasoning chain and answer. This method requires $b + 1$ forward passes, each producing more “thinking” tokens. 

\noindent \textbf{\textcircled{6} Self-Refinement} \cite{selfrefine} begins with a single CoT reasoning chain (\textcircled{1}) and then sequentially prompts the LVLM to refine and improve it for $k$ rounds. This process enables the LVLM to self-correct and self-verify over successive iterations. This method requires $k + 1$ forward passes, the first producing more “thinking” tokens and the successive passes producing more "self-reflective" tokens. 

\noindent \textbf{\textcircled{7} Describe-Answer} \cite{Guo2022FromIT} first prompts the LVLM to caption the image $I$ in full detail, including the objects present and their interactions. The generated caption, together with the image, question and a post-prompt, is then provided as input to the LVLM, which is asked to produce the final answer. This checks whether the LVLM’s performance improves if elicited with a text form of the image. This method requires two forward passes, and only the first pass requires more "thinking tokens" where those tokens are represented by the caption.  

\noindent \textbf{\textcircled{8} Compositional Chain-of-Thought} prompting \cite{ccot} first prompts the LVLM to generate a scene graph for the image $I$ given the \texttt{question}. The generated scene graph, together with the image, question and a post-prompt, is then provided as input to the LVLM, which is asked to produce the final answer. Similar to \textcircled{6}, this method requires two forward passes, and only the first pass requires more "thinking tokens" where those tokens are represented by the scene graph.  

\noindent \textbf{\textcircled{9} Prompt Repetition} is a recent technique \cite{Leviathan2025PromptRI} that simply repeats the \texttt{question}. In this setting, \texttt{think-prompt} = \texttt{question}. Since LVLMs are trained and evaluated autoregressively, earlier prompt tokens cannot attend to later ones. Repeating the question twice helps address this limitation, creating an effect of bidirectional attention. This technique has also proven valuable for language embeddings \cite{springer2025repetition}. This method uses a single forward pass, without producing any "thinking tokens". 

\section{Experiments and Takeaways}

\begin{table*}[p]
\centering
\caption{Test-Time scaling results on the Qwen series, with \colorbox{red!25}{degradation} means a lower score than baseline, and \colorbox{green!25}{improvement} means a higher score than baseline}
\setlength{\tabcolsep}{4pt}
\renewcommand{\arraystretch}{1.15}
\scriptsize
\scalebox{0.8}{
\begin{tabular}{c l *{5}{c} |c}
\toprule
\multirow{2}{*}{\rotatebox[origin=c]{90}{Model}} & \multirow{2}{*}{Method} & \multicolumn{6}{c}{Dataset} \\
\cmidrule(lr){3-8}
 &  & MMStar & RealWorldQA & HallusionBench & WeMath & LogicVista & Runtime (s) \\
\midrule
\multirow{9}{*}{\rotatebox[origin=c]{90}{\textbf{Qwen3-VL-2B}}} & Baseline \cite{Qwen3-VL} & 51.69 & 65.10 & 63.72 & 35.46 & 32.59  & 0.1 \\
 & CoT \cite{wei2022chain} & \cellcolor{green!55}{61.14} & \cellcolor{red!36}{60.65} & \cellcolor{green!66}{69.93} & \cellcolor{green!68}{55.92} & \cellcolor{green!20}{37.95} & 4.5 \\
 & S-CoT & \cellcolor{green!22}{55.05} & \cellcolor{red!80}{52.94} & \cellcolor{green!52}{68.35} & \cellcolor{green!55}{51.03} & \cellcolor{green!37}{40.40} & 4.6 \\
 & PaS \cite{plansolve} & \cellcolor{green!40}{57.33} & \cellcolor{red!26}{63.27} & \cellcolor{green!73}{71.50} & \cellcolor{green!68}{55.40} & \cellcolor{green!37}{40.62} & 4.4 \\
 & Self-Consistency \cite{selfconsistency} & \cellcolor{green!80}{63.83} & \cellcolor{red!15}{64.31} & \cellcolor{green!73}{71.08} & \cellcolor{green!80}{64.25} & \cellcolor{green!46}{46.88} & 40.2 \\
 & Self-Refinement \cite{selfrefine} & \cellcolor{green!54}{60.87} & \cellcolor{red!44}{59.22} & \cellcolor{green!64}{69.72} & \cellcolor{green!67}{56.15} & \cellcolor{green!21}{39.06} & 9.0 \\
 & Self-Aggregation \cite{selaggregate} & \cellcolor{green!67}{62.26} & \cellcolor{red!30}{61.70} & \cellcolor{green!76}{71.50} & \cellcolor{green!75}{58.68} & \cellcolor{green!41}{41.07} & 24.0 \\
 & Describe-Answer \cite{Guo2022FromIT} & \cellcolor{green!53}{53.49} & \cellcolor{green!18}{66.54} & \cellcolor{green!34}{66.35} & \cellcolor{green!17}{45.00} & \cellcolor{red!80}{30.13} & 6.4 \\
 & CCoT \cite{ccot} & \cellcolor{green!16}{52.65} & \cellcolor{red!29}{61.96} & \cellcolor{red!38}{60.57} & \cellcolor{green!70}{48.74} & \cellcolor{green!23}{34.60} & 5.8 \\
 & Prompt Repetition \cite{Leviathan2025PromptRI} & \cellcolor{red!20}{50.62} & \cellcolor{green!11}{65.36} & \cellcolor{green!13}{64.04} & \cellcolor{green!23}{39.20} & 32.59 & 0.1 \\
\midrule
\multirow{9}{*}{\rotatebox[origin=c]{90}{\textbf{Qwen3-VL-4B}}} & Baseline \cite{Qwen3-VL} & 61.71 & 71.76 & 70.66 & 58.22 & 40.85 & 0.1 \\
 & CoT \cite{wei2022chain} & \cellcolor{green!43}{68.59} & \cellcolor{red!28}{69.80} & \cellcolor{green!57}{75.50} & \cellcolor{green!55}{71.55} & \cellcolor{green!54}{57.59} & 5.9 \\
 & S-CoT & \cellcolor{green!49}{68.47} & \cellcolor{red!80}{64.18} & \cellcolor{green!57}{75.50} & \cellcolor{green!52}{70.86} & \cellcolor{green!52}{56.70} & 8.2 \\
 & PaS & \cellcolor{green!48}{69.38} & \cellcolor{red!19}{70.46} & \cellcolor{green!48}{74.76} & \cellcolor{green!38}{67.41} & \cellcolor{green!45}{54.69} & 8.0 \\ 
 & Self-Consistency \cite{selfconsistency} & \cellcolor{green!71}{71.07} & \cellcolor{green!18}{72.55} & \cellcolor{green!49}{74.97} & \cellcolor{green!64}{74.25} & \cellcolor{green!72}{62.95} & 44.8 \\
 & Self-Refinement \cite{selfrefine} & \cellcolor{green!39}{68.19} & \cellcolor{red!41}{67.58} & \cellcolor{green!36}{73.29} & \cellcolor{green!38}{68.33} & \cellcolor{green!59}{58.71} & 11.2 \\
 & Self-Aggregation \cite{selaggregate} & \cellcolor{green!71}{70.51} & \cellcolor{red!17}{70.98} & \cellcolor{green!60}{75.71} & \cellcolor{green!69}{75.40} & \cellcolor{green!64}{60.94} & 31.4 \\
 & Describe-Answer \cite{Guo2022FromIT} & \cellcolor{green!13}{62.62} & \cellcolor{red!24}{70.59} & \cellcolor{green!19}{71.29} & \cellcolor{red!13}{56.95} & \cellcolor{green!15}{47.32} & 9.0 \\
 & CCoT \cite{ccot} & \cellcolor{green!22}{62.81} & \cellcolor{red!34}{68.76} & \cellcolor{red!11}{69.61} & \cellcolor{green!14}{57.59} & \cellcolor{green!11}{41.07} & 4.7 \\
 & Prompt Repetition \cite{Leviathan2025PromptRI} & \cellcolor{green!13}{62.03} & \cellcolor{red!14}{71.11} & \cellcolor{red!13}{70.45} & \cellcolor{red!15}{54.37} & \cellcolor{green!16}{43.53} & 0.1 \\
\midrule
\multirow{9}{*}{\rotatebox[origin=c]{90}{\textbf{Qwen3-VL-8B}}} & Baseline \cite{Qwen3-VL} & 64.95 & 69.54 & 72.98 & 57.76 & 40.62 & 0.1 \\
 & CoT \cite{wei2022chain} & \cellcolor{green!44}{69.77} & \cellcolor{green!51}{72.03} & \cellcolor{green!55}{76.55} & \cellcolor{green!55}{74.37} & \cellcolor{green!43}{55.13} & 5.9 \\
 & S-CoT & \cellcolor{green!38}{69.08} & \cellcolor{red!14}{68.63} & \cellcolor{green!35}{75.08} & \cellcolor{green!52}{73.62} & \cellcolor{green!48}{56.70} & 7.2 \\
 & PaS & \cellcolor{green!50}{70.47} & \cellcolor{red!5}{69.28} & \cellcolor{green!36}{75.29} & \cellcolor{green!57}{74.89} & \cellcolor{green!51}{58.04} & 6.9 \\ 
 & Self-Consistency \cite{selfconsistency} & \cellcolor{green!80}{72.99} & \cellcolor{green!80}{72.68} & \cellcolor{green!44}{75.71} & \cellcolor{green!70}{76.90} & \cellcolor{green!65}{61.61} & 44.8 \\
 & Self-Refinement \cite{selfrefine} & \cellcolor{green!40}{69.16} & \cellcolor{red!14}{68.63} & 72.98 & \cellcolor{green!47}{72.41} & \cellcolor{green!49}{56.92} & 11.3 \\
 & Self-Aggregation \cite{selaggregate} & \cellcolor{green!70}{72.19} & \cellcolor{green!58}{71.11} & \cellcolor{green!38}{75.50} & \cellcolor{green!72}{77.47} & \cellcolor{green!80}{63.62} & 31.1 \\
 & Describe-Answer \cite{Guo2022FromIT} & \cellcolor{red!25}{63.92} & \cellcolor{red!21}{69.02} & 72.98 & \cellcolor{red!15}{56.61} & \cellcolor{red!15}{37.72} & 9.6 \\
 & CCoT \cite{wei2022chain} & \cellcolor{green!18}{65.47} & \cellcolor{red!21}{69.02} & \cellcolor{red!15}{72.24} & \cellcolor{green!20}{60.40} & \cellcolor{red!22}{38.17} & 5.9 \\
 & Prompt Repetition \cite{Leviathan2025PromptRI} & \cellcolor{green!11}{65.02} & \cellcolor{green!11}{69.80} & \cellcolor{red!11}{72.45} & \cellcolor{green!12}{58.22} & 40.62 & 0.1 \\
\midrule
\multirow{9}{*}{\rotatebox[origin=c]{90}{\textbf{Qwen2.5-VL-7B}}} & Baseline \cite{Qwen3-VL} & 62.57 & 70.20 & 72.03 & 58.28 & 39.06 & 0.1 \\
 & CoT \cite{wei2022chain} & \cellcolor{green!24}{63.46} & \cellcolor{red!32}{65.88} & \cellcolor{red!35}{69.61} & \cellcolor{green!33}{65.40} & \cellcolor{red!13}{38.17} & 2.1 \\
 & S-CoT & \cellcolor{green!37}{64.11} & \cellcolor{red!26}{61.70} & \cellcolor{red!26}{70.24} & \cellcolor{green!18}{62.24} & \cellcolor{green!29}{42.41} & 2.7 \\
 & PaS & \cellcolor{red!19}{61.88} & \cellcolor{red!45}{64.18} & \cellcolor{red!39}{69.30} & \cellcolor{green!29}{64.48} & \cellcolor{green!78}{44.42} & 2.5 \\ 
 & Self-Consistency \cite{selfconsistency} & \cellcolor{green!62}{65.29} & \cellcolor{red!32}{66.01} & \cellcolor{red!20}{71.08} & \cellcolor{green!47}{69.25} & \cellcolor{green!23}{41.52} & 17.3 \\
 & Self-Refinement \cite{selfrefine} & \cellcolor{red!25}{61.33} & \cellcolor{red!80}{47.97} & \cellcolor{red!80}{68.66} & \cellcolor{green!26}{63.85} & \cellcolor{green!12}{41.07} & 4.6 \\
 & Self-Aggregation \cite{selaggregate} & \cellcolor{red!12}{62.04} & \cellcolor{red!28}{61.44} & \cellcolor{red!23}{70.45} & \cellcolor{green!39}{67.59} & \cellcolor{green!80}{49.11} & 11.8 \\
 & Describe-Answer \cite{Guo2022FromIT} & \cellcolor{red!49}{60.41} & \cellcolor{red!30}{67.45} & \cellcolor{red!23}{70.03} & \cellcolor{red!25}{54.71} & \cellcolor{red!24}{36.16} & 5.0 \\
 & CCoT \cite{wei2022chain} & \cellcolor{red!44}{60.58} & \cellcolor{red!24}{62.75} & \cellcolor{red!37}{69.51} & \cellcolor{green!14}{59.25} & \cellcolor{red!11}{38.39} & 3.3 \\
 & Prompt Repetition \cite{Leviathan2025PromptRI} & \cellcolor{red!48}{60.38} & \cellcolor{red!30}{67.97} & \cellcolor{red!16}{71.29} & \cellcolor{green!15}{59.48} & \cellcolor{green!19}{40.62} & 0.1 \\
 \midrule
\multirow{9}{*}{\rotatebox[origin=c]{90}{\textbf{Qwen3-VL-32B}}} & Baseline \cite{Qwen3-VL} & 72.66 & 77.25 & 74.76 & 66.03 & 44.42 & 0.2 \\
 & CoT \cite{wei2022chain} & \cellcolor{green!46}{74.88} & \cellcolor{red!31}{74.64} & \cellcolor{green!45}{78.13} & \cellcolor{green!54}{79.43} & \cellcolor{green!52}{66.52} & 12.2 \\
 & S-CoT & \cellcolor{green!43}{74.67} & \cellcolor{red!11}{76.34} & \cellcolor{green!18}{76.55} & \cellcolor{green!52}{78.68} & \cellcolor{green!52}{66.52} & 19.5 \\
 & PaS & \cellcolor{green!46}{76.05} & \cellcolor{red!31}{76.34} & \cellcolor{green!45}{76.13} & \cellcolor{green!54}{78.97} & \cellcolor{green!52}{66.52} & 18.9 \\
 & Self-Consistency \cite{selfconsistency} & \cellcolor{green!80}{76.46} & \cellcolor{green!22}{77.91} & \cellcolor{green!35}{77.71} & \cellcolor{green!60}{80.40} & \cellcolor{green!59}{68.75} & 88.9 \\
 & Self-Refinement \cite{selfrefine} & \cellcolor{green!21}{73.21} & \cellcolor{red!80}{71.50} & \cellcolor{green!40}{77.92} & \cellcolor{green!57}{80.17} & \cellcolor{green!56}{67.41} & 26.5 \\
 & Self-Aggregation \cite{selaggregate} & \cellcolor{green!70}{75.85} & \cellcolor{red!20}{75.56} & \cellcolor{green!31}{77.50} & \cellcolor{green!80}{81.44} & \cellcolor{green!63}{70.76} & 65.6 \\
 & Describe-Answer \cite{Guo2022FromIT} & \cellcolor{red!80}{70.20} & \cellcolor{red!11}{76.34} & \cellcolor{red!14}{74.45} & \cellcolor{red!18}{64.48} & \cellcolor{green!14}{45.76} & 26.9 \\
 & CCoT \cite{wei2022chain} & \cellcolor{red!13}{72.36} & \cellcolor{red!51}{73.73} & \cellcolor{red!12}{74.55} & \cellcolor{green!29}{71.67} & \cellcolor{green!19}{47.54} & 17.5 \\
 & Prompt Repetition \cite{Leviathan2025PromptRI} & \cellcolor{red!19}{72.07} & \cellcolor{green!80}{79.08} & \cellcolor{green!14}{75.08} & \cellcolor{green!11}{66.15} & \cellcolor{red!12}{43.75} & 0.2 \\
\midrule
\multirow{9}{*}{\rotatebox[origin=c]{90}{\textbf{Qwen2.5-VL-32B}}} & Baseline \cite{Qwen2.5-VL} & 67.35 & 71.24 & 70.35 & 65.80 & 44.87 & 0.2 \\
 & CoT \cite{wei2022chain} & \cellcolor{red!22}{66.87} & \cellcolor{red!26}{66.80} & \cellcolor{green!14}{70.77} & \cellcolor{green!42}{74.31} & \cellcolor{green!44}{56.03} & 8.7 \\
 & S-CoT & \cellcolor{green!57}{68.37} & \cellcolor{red!13}{69.02} & \cellcolor{green!32}{71.92} & \cellcolor{green!37}{73.28} & \cellcolor{green!46}{56.47} & 15.2 \\
 & PaS & \cellcolor{green!55}{68.31} & \cellcolor{red!24}{69.28} & \cellcolor{green!12}{70.45} & \cellcolor{green!42}{74.48} & \cellcolor{green!44}{56.03} & 14.8\\
 & Self-Consistency \cite{selfconsistency} & \cellcolor{green!61}{68.97} & \cellcolor{red!19}{68.10} & \cellcolor{green!26}{71.50} & \cellcolor{green!60}{76.78} & \cellcolor{green!48}{56.47} & 64.1 \\
 & Self-Refinement \cite{selfrefine} & \cellcolor{red!80}{65.62} & \cellcolor{red!80}{49.93} & \cellcolor{green!24}{71.40} & \cellcolor{green!32}{72.87} & \cellcolor{green!40}{54.91} & 20.6 \\
 & Self-Aggregation \cite{selaggregate} & \cellcolor{red!15}{67.16} & \cellcolor{red!44}{60.78} & \cellcolor{green!13}{70.45} & \cellcolor{green!55}{76.21} & \cellcolor{green!52}{58.04} & 48.8 \\
 & Describe-Answer \cite{Guo2022FromIT} & \cellcolor{red!32}{66.32} & \cellcolor{red!16}{69.15} & \cellcolor{red!14}{70.14} & \cellcolor{red!13}{63.05} & \cellcolor{red!21}{39.51} & 23.1 \\
 & CCoT \cite{wei2022chain} & \cellcolor{red!76}{65.58} & \cellcolor{red!18}{67.32} & \cellcolor{green!80}{73.50} & \cellcolor{green!23}{70.29} & \cellcolor{green!19}{46.43} & 21.0 \\
 & Prompt Repetition \cite{Leviathan2025PromptRI} & \cellcolor{red!22}{66.88} & \cellcolor{red!14}{69.41} & \cellcolor{green!13}{70.45} & \cellcolor{green!11}{66.26} & \cellcolor{red!11}{43.75} & 0.2 \\
\midrule
\multirow{9}{*}{\rotatebox[origin=c]{90}{\textbf{Qwen2.5-VL-72B}}} & Baseline \cite{Qwen2.5-VL} & 68.65 & 69.67 & 74.97 & 72.99 & 47.54 & 0.3 \\
 & CoT \cite{wei2022chain} & \cellcolor{red!12}{68.58} & \cellcolor{red!20}{67.58} & \cellcolor{red!52}{72.98} & \cellcolor{green!12}{73.05} & \cellcolor{green!41}{56.03} & 11.6 \\
 & S-CoT & \cellcolor{green!14}{68.77} & \cellcolor{red!13}{68.50} & \cellcolor{red!30}{73.82} & 72.99 & \cellcolor{green!80}{59.82} & 16.3 \\
 & Self-Consistency \cite{selfconsistency} & \cellcolor{green!80}{70.86} & \cellcolor{green!35}{71.37} & \cellcolor{red!52}{72.98} & \cellcolor{green!80}{74.66} & \cellcolor{green!73}{58.93} & 85.7 \\
 & Self-Refinement \cite{selfrefine} & \cellcolor{green!17}{68.89} & \cellcolor{red!80}{49.67} & \cellcolor{red!80}{71.61} & \cellcolor{red!80}{69.37} & \cellcolor{green!46}{56.70} & 25.3 \\
 & Self-Aggregation \cite{selaggregate} & \cellcolor{green!70}{70.44} & \cellcolor{red!34}{63.14} & \cellcolor{red!59}{72.87} & \cellcolor{green!51}{73.97} & \cellcolor{green!56}{57.14} & 68.6 \\
 & Describe-Answer \cite{Guo2022FromIT} & \cellcolor{red!80}{66.78} & \cellcolor{red!24}{67.19} & \cellcolor{red!67}{72.66} & \cellcolor{red!42}{71.55} & \cellcolor{red!18}{44.42} & 26.7 \\
 & CCoT \cite{wei2022chain} & \cellcolor{red!32}{67.99} & \cellcolor{red!11}{69.41} & \cellcolor{red!37}{73.40} & \cellcolor{red!42}{71.55} & \cellcolor{green!27}{50.67} & 15.6 \\
 & Prompt Repetition \cite{Leviathan2025PromptRI} & \cellcolor{green!36}{69.40} & \cellcolor{red!14}{69.02} & \cellcolor{green!15}{75.18} & \cellcolor{red!62}{71.84} & \cellcolor{red!12}{45.98} & 0.2 \\
\bottomrule
\end{tabular}
}
\label{tab:qwen_methods_datasets}
\end{table*}

\noindent \textbf{Models:} In this work, we study 13 \textit{instruction-tuned} models spanning a range of scales from the Qwen-2.5-VL \cite{Qwen2.5-VL}, Qwen-3-VL \cite{Qwen3-VL} series, InternVL-3.5 \cite{Wang2025InternVL35AO} series and the Molmo2 \cite{Clark2026Molmo2OW} series. For Qwen-2.5-VL, we consider model sizes of \{7B, 32B and 72B\}. For Qwen-3-VL, we consider model sizes of \{2B, 4B, 8B, and 32B\}. For InternVL-3.5, we consider model sizes of \{2B, 4B, 8B, and 38B\}. Finally, for Molmo2, we consider model sizes of \{4B, 8B\}. To the best of our knowledge, the Qwen-VL and InternVL series are the strongest open-source models, achieving close-to and even surpassing top-tier closed-source models on various benchmarks \cite{Qwen2.5-VL, Qwen3-VL}, and heavily used in industry. For all models, we set the maximum token length to 1,024. The total cost of our experiments is approximately \$6,000.  
\newline
\newline
\noindent \textbf{Benchmarks:} We consider six diverse benchmarks: MMStar \cite{mmstar}, RealWorldQA \cite{realworldqa}, HallusionBench \cite{hallusionbench}, WeMath \cite{wemath}, LogicVista \cite{logicvista} and A-OKVQA \cite{Schwenk2022AOKVQAAB}. MMStar assesses coarse perception, fine-grained perception, mathematics, science and technology, instance reasoning, and logical reasoning. It therefore serves as both a perception and a reasoning benchmark, and consists of curated, reviewed samples drawn from existing benchmarks such as MMBench \cite{liu2024mmbench}, MathVista \cite{lu2024mathvista}, MMMU \cite{yue2023mmmu}, ScienceQA \cite{lu2022learn}, SEED \cite{li2023seed}, and AI2D \cite{ai2d}. RealWorldQA is a perception benchmark focused on spatial understanding. A-OKVQA is also a perception benchmark. HallusionBench is both a perception and visual reasoning benchmark, designed to evaluate hallucination in LVLMs. WeMath and LogicVista are mathematical and logical reasoning benchmarks, respectively.
\newline
\newline
\noindent \textbf{Results:} Results are presented in Table \ref{tab:qwen_methods_datasets} for the Qwen Series. We color-code metrics in red when results degrade relative to the baseline, and in green otherwise, with the color intensity indicating the magnitude of the improvement or degradation. We also report the runtime (in seconds) in the last column for each method and model. Runtimes are measured on a single NVIDIA H200 GPU with 140 GB of VRAM, except for Qwen-2.5-VL-72B, which was run in parallel across two H200 GPUs due to memory constrains of a single H200 GPU. For each benchmark, we compute the runtime by averaging over all samples in that benchmark. We then obtain a single runtime value per method and model by averaging across all five benchmarks. By analyzing the results across all models and benchmarks, we find that overall, test-time scaling greatly improves performance across most models and benchmarks. 

\begin{figure}
    \centering
    \includegraphics[width=1\linewidth]{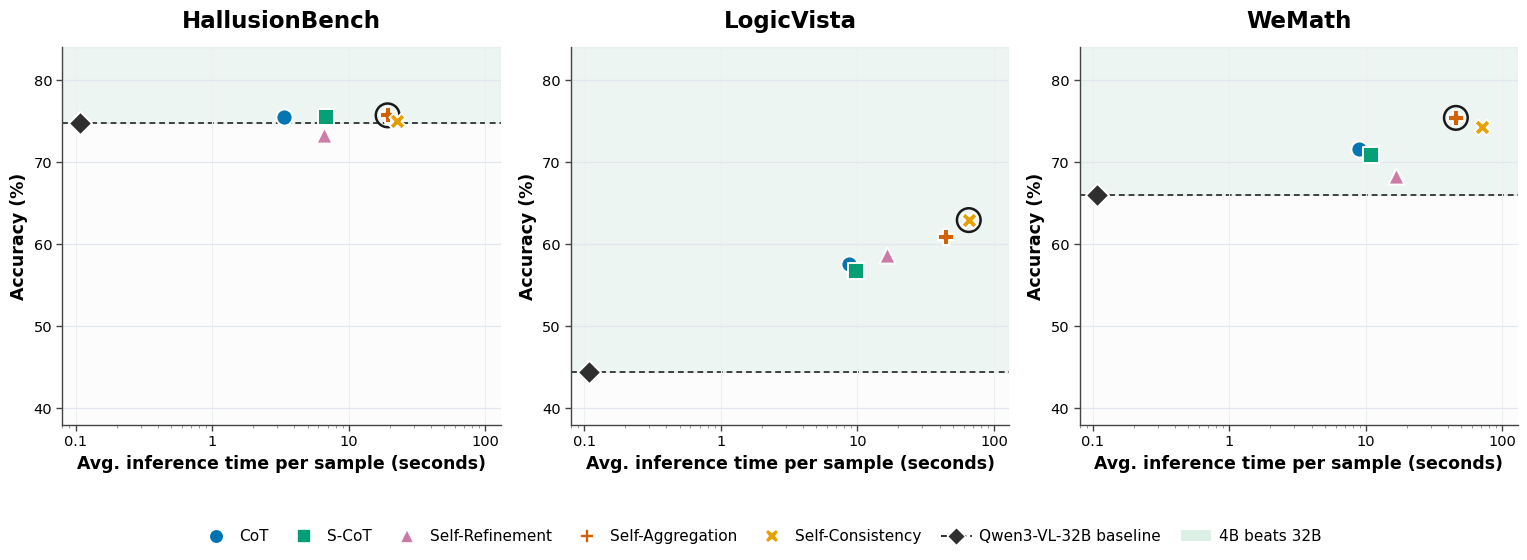}
    \caption{Compute vs Accuracy Pareto frontiers. We circle the best method. }
    \label{fig:pareto_frontier}
\end{figure}

\noindent Interestingly, and different from prior LLM works which report that small models do not benefit from test-time scaling \cite{Lanham2023MeasuringFI, Ou2025EmpoweringLM, magister-etal-2023-teaching}, we observe the opposite in LVLMs. Our results suggest that small models benefit the most from test-time scaling. For example, self-consistency boosts the performance of Qwen3-VL-2B from 35\% to 64\% on WeMath (approximately a 30\% increase in performance). Our findings also suggest that small models with test-time scaling can outperform larger ones. For example, Qwen3-VL-4B with simple CoT and S-CoT already surpasses the baseline performance of Qwen3-VL-32B on HallusionBench, WeMath and LogicVista. On LogicVista for example, Qwen3-VL-4B with CoT achives a +13\% improvement over Qwen3-VL-32B. The gain is even larger with Self-Consistency. We also find that simple prompt repetition helps on tasks that require reasoning across the prompt, such as when the prompt describes a mathematical problem and earlier prompt tokens benefit from future prompt tokens. For example, on WeMath, simple prompt repetition yields an improvement of about +4\% for Qwen3-VL-2B, while the gains are less pronounced for other models. Figure \ref{fig:pareto_frontier} presents the Pareto frontiers for HallusionBench, LogicVista, and WeMath, comparing Qwen3-VL-4B with test-time scaling  against the $\times$8 larger Qwen3-VL-32B baseline ($\blacklozenge$, dashed line). The x-axis represents compute time, while the y-axis shows accuracy. The figure illustrates whether, under a fixed compute budget, it is better to allocate resources to a smaller model with test-time scaling or to a larger model without it. For example, at a compute budget of approximately 9 seconds, the 4B model matches the performance of the 32B model on HallusionBench, nearly matches it on MMStar, and outperforms it on both LogicVista and WeMath. Overall, we find that simple Chain-of-Thought prompting (CoT) offers the best tradeoff between runtime and performance improvements. 

\noindent It is important to note that a very recent work \cite{kaya2026efficient} also studies some test-time scaling methods for a small vision-language model (SmolVLM2 \cite{marafioti2025smolvlm}) and report failures, and other works \cite{ccot} investigate chain-of-thought prompting for the LLaVa series, and similarly finds it ineffective. To understand this discrepancy between our findings and the literature, we examined model behavior and found that LLaVA-OneVision \cite{li2025llavaonevision} and SmolVLM2 \cite{marafioti2025smolvlm} often show poor instruction compliance: they answer questions directly while ignoring the \texttt{think-prompt}. They behaved more like specialized VQA systems rather than instruction-following chat agents (see Section \ref{supp:model_failures} of the Supp. material for examples). Therefore, our results suggest that \textit{capable} instruction-following models benefit the most from test-time scaling. This suggests that we can deploy cheaper, faster, lower-memory VLMs and use test-time scaling to get near--large-model accuracy (and even outperform them) in production and research. 

\Finding{1}{Smaller models benefit the most}{%
Small (2B to 8B scale), yet capable instruction-tuned models, benefit the most from test-time scaling, often reaching close to performance of larger models (and sometimes even outperforming them), suggesting an accuracy--compute trade-off where cheaper, faster and lower-memory VLMs can be deployed.}  

\noindent Small models may not have the same direct ability to solve complex tasks as larger models. However, they still contain relevant information compressed in their weights and possess some capacity for reasoning. When this reasoning ability is activated, they can decompose a difficult problem into a sequence of simpler steps, solve each step one by one, and ultimately arrive at the correct answer.
\newline
\newline
\noindent We also find that test-time scaling methods often degrade performance (rather than maintaining it) on primarily perception-focused benchmarks that require limited or no reasoning (e.g., RealWorldQA). We additionally verify this claim in two ways: (1) by reporting additional results on another perception-only benchmark, A-OKVQA (Section \ref{supp:aokvqa} of the Supp. material), and (2) by dissecting the individual category scores on MMStar, showing that the degradation occurs mainly in the non-reasoning categories (Section \ref{supp:mmstar_categorical_perception_degrade} of the Supp. material). Notably, perception benchmarks are equally important, as they often reflect typical-day user queries. To test whether this effect also holds for strong closed-source models, we ran experiments on MMStar and RealWorldQA using GPT-5.2, OpenAI’s strongest model at the time of writing. The results are reported in Section \ref{supp:gpt5.2} of the Supp. material. We find that this issue is less pronounced for GPT-5.2, but in some cases, still causing degradation. 

\begin{figure}
    \centering
    \includegraphics[width=1\linewidth]{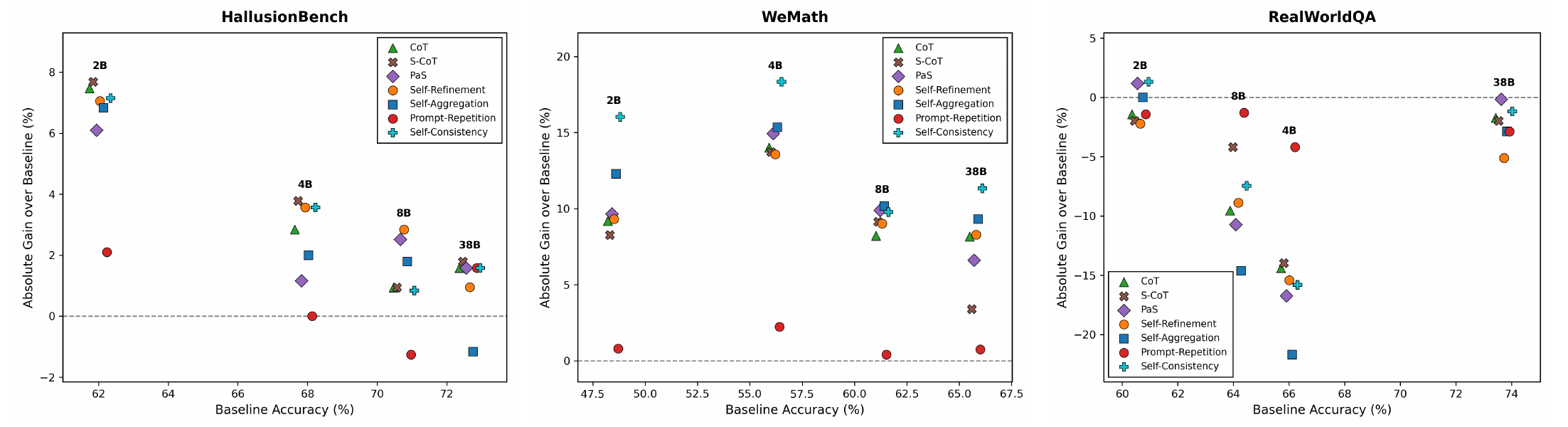}
    \caption{Absolute Gains of Test-Time Scaling Methods on the InternVL-3.5 series}
    \label{fig:intenrvl}
\end{figure}

\noindent \textbf{Generalization to other LVLMs:} Figure \ref{fig:intenrvl} shows the absolute gains (in \%) of test-time scaling methods over the baseline (dashed line) for the InternVL-3.5 series. The results confirm our main findings: smaller models benefit most from test-time scaling, while test-time scaling degrades performance on perceptual-only benchmarks such as RealWorldQA. Tabular results are in Section \ref{supp:intnervl} of the Supp. Material for InternVL-3.5 and in Section \ref{supp:molmo2} for the Molmo2 series. 
\newline
\newline
\noindent A question that remains unanswered is: \textit{why does extra test-time compute harm perception-only benchmarks?} We answer this via Token Budget analysis.

\noindent \textbf{Token Budget:} Token Budget refers to the maximum number of test-time scaling tokens generated. To study how token length affects performance gains from test-time scaling methods in vision-language models, we repeated all experiments (Section \ref{section:method}) using a reduced token budget. Specifically, we set the maximum output length to 300 tokens. Because we use the reliable answer extraction method (Section \ref{section:method}), we can still reliably extract the final answer even if the reasoning chain is truncated at 300 tokens. Results are presented in Table \ref{tab:token_budget}. We report the difference in score when the reasoning chain is stopped early at 300 tokens. 
\newline
\newline
In contrast to findings in the LLM literature suggesting that models often “know” the answer before producing a chain of thought \cite{Lanham2023MeasuringFI}—that is, the conclusion is reached early and the final answer may not depend on the reasoning trace—our experiments indicate the opposite. Reducing the token budget degrades performance, implying (1) the reasoning chain drives the correct answer, and (2) using more tokens to generate intermediate steps is beneficial for reaching the correct answer. This generally holds across all reasoning benchmarks, but not across perceptual and hallucination benchmarks. On RealWorldQA and often on HallusionBench (as seen in Table \ref{tab:token_budget}), increasing the token budget can hurt performance, or at least does not result in any improvements. In Section \ref{supp:token_cutoff} of the Supp. material, we additionally verify this also on (1) the perceptual categories of MMStar, (2) A-OKVQA dataset and (3) InternVL-3.5. This suggests that short answers are best for perceptual tasks, and overly verbose responses drive the LVLM to "lose focus"; if the LVLM is given the opportunity to overthink (via more tokens), it tends to elaborate plausible narratives, make wrong assumptions and accumulate chain errors, leading to hallucinations and eventually drifting away from the correct answer. We provide supporting evidence for this hypothesis via internal model attention analysis in the next section. Overall, our analysis suggests that test-time scaling is primarily helpful when it enables genuinely useful reasoning, For tasks that do not require reasoning, applying test-time scaling yields generally no improvement and, in the worst case, even degrade performance. This implies that \textbf{substantial compute can be saved} by identifying whether a task actually requires test-time scaling, for example by training a simple binary classifier. 

\begin{table*}[!t]
\centering
\caption{Early stopping at token 300. We report $\Delta = \text{score}_{300} - \text{score}_{1024}$. \colorbox{green!25}{\strut improvement} indicates a positive $\Delta$ (300-token early stopping improves performance). \colorbox{red!25}{degradation} indicates negative $\Delta$ (performance drops under the 300-token stopping).}
\setlength{\tabcolsep}{3.2pt}
\renewcommand{\arraystretch}{1.08}
\small
\label{tab:delta_300_1024_rearranged}

\scalebox{0.8}{%
\begin{tabular}{l l c c c c c}
\toprule
\textbf{Model} & \textbf{Method} & \textbf{MMStar} & \textbf{RealWorldQA} & \textbf{Hall.Bench} & \textbf{WeMath} & \textbf{LogicVista} \\
\midrule

\multirow{5}{*}{Qwen3-VL-4B}
& CoT       & \cellcolor{red!35}  -2.07 & \cellcolor{green!25} 0.79 & \cellcolor{red!15}  -0.42 & \cellcolor{red!65}  -11.32 & \cellcolor{red!55}  -7.37 \\
& S-CoT       & \cellcolor{red!35}  -2.80 & \cellcolor{green!45} 3.79 & \cellcolor{red!15}  -0.11 & \cellcolor{red!65}  -14.89 & \cellcolor{red!65}  -13.61 \\
& Self-Ref. & \cellcolor{red!35}  -2.31 & \cellcolor{red!15}  -0.26 & \cellcolor{red!15}  -0.31 & \cellcolor{red!55}  -7.35  & \cellcolor{red!35}  -2.45 \\
& Self-Agg. & \cellcolor{red!35}  -2.57 & \cellcolor{green!15} 0.26 & \cellcolor{red!25}  -0.74 & \cellcolor{red!65}  -15.06 & \cellcolor{red!65}  -13.84 \\
& Self-Cons.& \cellcolor{red!35}  -3.26 & \cellcolor{green!25} 0.65 & \cellcolor{green!25} 1.06 & \cellcolor{red!65}  -10.46 & \cellcolor{red!65}  -9.37 \\
\midrule

\multirow{5}{*}{Qwen3-VL-8B}
& CoT       & \cellcolor{red!25}  -0.55 & \cellcolor{red!25}  -0.79 & \cellcolor{red!35}  -2.10 & \cellcolor{red!55}  -6.04  & \cellcolor{red!15}  -0.44 \\
& S-CoT       & \cellcolor{red!25}  -0.80 & \cellcolor{green!25} 0.52 & \cellcolor{green!25} 0.84 & \cellcolor{red!65}  -10.29 & \cellcolor{red!35}  -1.57 \\
& Self-Ref. & \cellcolor{red!35}  -1.82 & \cellcolor{red!15}  -0.13 & \cellcolor{green!25} 0.84 & \cellcolor{red!45}  -5.92  & \cellcolor{red!25}  -1.34 \\
& Self-Agg. & \cellcolor{red!35}  -2.87 & \cellcolor{red!25}  -1.18 & \cellcolor{red!25}  -1.16 & \cellcolor{red!65}  -9.60  & \cellcolor{red!55}  -8.04 \\
& Self-Cons.& \cellcolor{red!45}  -3.15 & \cellcolor{red!25}  -0.78 & \cellcolor{red!15}  -0.32 & \cellcolor{red!55}  -7.88  & \cellcolor{red!35}  -1.57 \\
\midrule

\multirow{5}{*}{Qwen3-VL-32B}
& CoT       & \cellcolor{red!25}  -0.62 & 0.00 & \cellcolor{red!35}  -1.58 & \cellcolor{red!55}  -7.25  & \cellcolor{red!45}  -3.13 \\
& S-CoT       & \cellcolor{red!25}  -1.37 & \cellcolor{green!35} 1.57 & \cellcolor{green!15} 0.21 & \cellcolor{red!65}  -11.84 & \cellcolor{red!45}  -5.58 \\
& Self-Ref. & \cellcolor{red!15}  -0.15 & \cellcolor{green!15} 0.13 & \cellcolor{red!35}  -2.32 & \cellcolor{red!55}  -8.91  & \cellcolor{red!45}  -3.79 \\
& Self-Agg. & \cellcolor{red!25}  -1.36 & \cellcolor{green!25} 0.65 & \cellcolor{red!25}  -0.63 & \cellcolor{red!65}  -10.23 & \cellcolor{red!45}  -4.91 \\
& Self-Cons.& \cellcolor{red!35}  -1.64 & \cellcolor{red!25}  -0.66 & \cellcolor{red!25}  -1.16 & \cellcolor{red!55}  -7.24  & \cellcolor{red!55}  -6.70 \\
\midrule

\multirow{5}{*}{Qwen2.5-VL-7B}
& CoT       & \cellcolor{red!25}  -0.58 & \cellcolor{green!15} 0.13 & \cellcolor{green!25} 1.37 & \cellcolor{red!35}  -2.24 & 0.00 \\
& S-CoT       & \cellcolor{red!25}  -1.11 & \cellcolor{green!15} 0.26 & \cellcolor{red!25}  -0.84 & \cellcolor{red!35}  -2.58 & 0.00 \\
& Self-Ref. & \cellcolor{red!25}  -0.57 & \cellcolor{green!15} 0.27 & \cellcolor{red!15}  -0.10 & \cellcolor{red!35}  -2.93 & \cellcolor{green!15} 0.22 \\
& Self-Agg. & \cellcolor{green!25} 1.24 & \cellcolor{green!25} 1.31 & \cellcolor{red!25}  -0.63 & \cellcolor{red!45}  -5.87 & \cellcolor{red!55}  -6.48 \\
& Self-Cons.& \cellcolor{green!15} 0.05 & \cellcolor{red!35}  -2.22 & \cellcolor{red!25}  -0.73 & \cellcolor{red!45}  -4.59 & \cellcolor{red!15}  -0.45 \\
\midrule

\multirow{5}{*}{Qwen2.5-VL-32B}
& CoT       & \cellcolor{red!25}  -1.28 & 0.00 & \cellcolor{green!25} 1.15 & \cellcolor{red!45}  -5.57 & \cellcolor{red!35}  -2.91 \\
& S-CoT       & \cellcolor{red!35}  -2.66 & \cellcolor{green!25} 0.52 & 0.00 & \cellcolor{red!65}  -9.26 & \cellcolor{red!55}  -7.59 \\
& Self-Ref. & \cellcolor{green!25} 0.63 & \cellcolor{green!35} 2.36 & \cellcolor{red!25}  -1.16 & \cellcolor{red!45}  -5.86 & \cellcolor{red!35}  -2.90 \\
& Self-Agg. & \cellcolor{green!15} 0.05 & \cellcolor{green!25} 0.66 & \cellcolor{green!25} 0.84 & \cellcolor{red!55}  -8.51 & \cellcolor{red!45}  -3.80 \\
& Self-Cons.& \cellcolor{red!15}  -0.48 & \cellcolor{green!15} 0.27 & \cellcolor{green!25} 0.95 & \cellcolor{red!55}  -7.30 & \cellcolor{green!25} 0.45 \\
\midrule

\multirow{5}{*}{Qwen2.5-VL-72B}
& CoT       & \cellcolor{red!15}  -0.34 & \cellcolor{green!35} 2.22 & \cellcolor{green!25} 0.94 & \cellcolor{green!15} 0.05 & 0.00 \\
& S-CoT       & \cellcolor{red!35}  -1.58 & \cellcolor{green!25} 1.04 & \cellcolor{red!25}  -0.53 & \cellcolor{red!25}  -0.75 & \cellcolor{red!15}  -0.44 \\
& Self-Ref. & \cellcolor{red!15}  -0.15 & \cellcolor{green!65} 10.07 & \cellcolor{green!35} 1.68 & \cellcolor{green!35} 2.76 & \cellcolor{red!15}  -0.23 \\
& Self-Agg. & \cellcolor{red!35}  -1.52 & \cellcolor{green!55} 6.14 & \cellcolor{green!35} 1.58 & \cellcolor{red!25}  -1.10 & \cellcolor{red!25}  -0.67 \\
\bottomrule
\end{tabular}%
}
\footnotesize
\label{tab:token_budget}
\end{table*}

\Finding{2}{LVLMs lose focus when given more compute than necessary}{%
Perception and hallucination benchmarks favor concise outputs, and extra test-time compute that enables verbosity often drives LVLMs to overthink and hallucinate, misleading the final answer. This suggests that a large amount of compute can be saved for tasks that do not require reasoning.}

\section{Attention Dynamics of Reasoning}
\begin{figure}[h]
    \centering
    \includegraphics[width=1\linewidth]{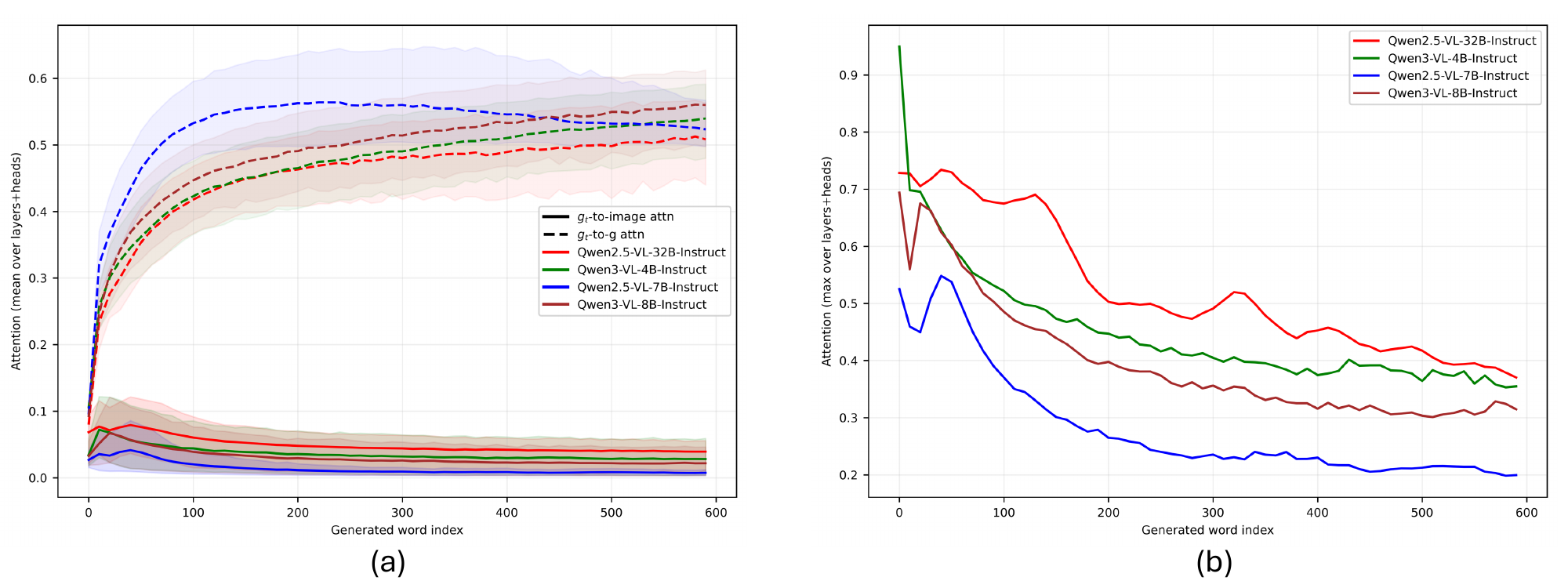}
    \caption{Attention dynamics during chain-of-thought generation. For each generated token $g_y$, we measure attention to image tokens $P_v^{ad}$ (\emph{solid curve}) and previously generated tokens (\emph{dashed curve}). \textbf{(a)} Average attention across layers, heads, and samples for WeMath. \textbf{(b)} Upper bound on image attention, defined as the maximum attention to image tokens across layers and heads.}
    \label{fig:attn}
\end{figure}

We remind the reader from Section \ref{section:method} that $P_v^{ad} \in \mathbb{R}^{N_m \times d}$ denotes the visual tokens produced by the adapter and passed to the LLM. In this section, we analyze how the LVLM’s chain-of-thought reasoning interacts with the image tokens $P_v^{ad}$.  Because most test-time scaling methods rely on CoT prompting as the underlying foundation, we adopt CoT as the basis for our analysis. Given an input prompt $P_t$ (Section \ref{section:method}), the LVLM generates a token sequence $g$. Let $g_y$ denote the $y$-th generated reasoning token. We examine, for each $g_y$, its interaction with (i) the image tokens $P_v^{ad}$ and (ii) the previously generated reasoning tokens $g_{0 \dots y-1}$. Concretely, at layer $l$ and attention head $h$, the attention map $a_l^h$ for token $g_y$ assigns attention over the $N_m$ image tokens ($P_v^{ad}$), the $T$ text prompt tokens ($P_t$), and the preceding generated tokens $g_{1 \dots y-1}$. We decompose $a_l^h$ for $g_y$ into three components: (1) attention allocation to the image tokens, $a^{img} = \sum a_l^h[1:N_m]$, (2) attention to the text prompt $a^{pmt} = \sum a[N_m+1: N_m+T]$, and (3) attention to previously generated reasoning tokens $a^{gen} = \sum a[N_m+T+1:N_m+T+(y-1)]$, where $y$ is the index of the current generated token. Figure \ref{fig:attn} plots $a^{img}$ and $a^{gen}$ attention traces as functions of the generated reasoning tokens for WeMath (Figure \ref{fig:attn}a) across different models, averaging over all layers and heads and over all samples in each dataset. The attention from $g_y$ to the $T$ text tokens in $P_t$ can be directly derived as $a^{pmt} = 1 - (\sum a^{img} + \sum a^{gen})$; because our focus is the contribution of image attention as tokens are generated, we omit this term from the plots. In Figure \ref{fig:attn}b, we additionally report an \textbf{upper bound} on image attention for WeMath, respectively, by taking, for each $g_y$, the maximum image-attention value across all layers and heads, i.e., we plot $max[a^{img}]$. Similar trends hold for other datasets and test-time scaling methods. We refer the reader to Section \ref{supp:img_attn} of the Supp. material for the corresponding results. Plotting details are provided in Section \ref{supp:attn_plot_details} of the Supp. material.

\begin{figure}[h]
    \centering
    \includegraphics[width=1\linewidth]{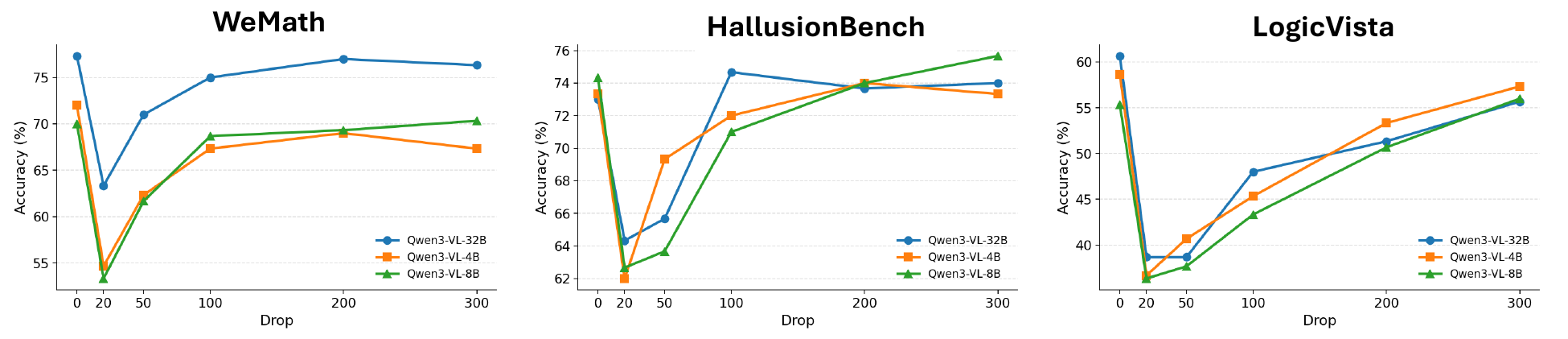}
    \caption{Image Token Dropping vs Accuracy at Different Generation Steps}
    \label{fig:dop_kv_cache}
\end{figure}

\noindent From Figure \ref{fig:attn}, we observe that across both datasets, attention mass quickly shifts toward previously generated reasoning tokens, while average attention to image tokens remains low and decays over the generation. Image attention peaks early and then drops rapidly, while attention to previously generated tokens increases and eventually dominates. This indicates that visual grounding is largely front-loaded: CoTs typically have an early, brief window of image-focused attention during which visual information is encoded and absorbed into the text tokens hidden representations. Even the maximum image attention per token drops steadily as generation progresses. 
\newline
\newline
\noindent Taken together, these patterns suggest that long CoTs rely increasingly on previously generated textual tokens for reasoning, and use primarily the visual information already encoded during that initial window rather than repeatedly attending to the image. This helps explain Takeaway 2: when test-time compute exceeds what is necessary and generation becomes overly verbose, the LVLM leans more heavily on the generated tokens to reason, producing assumptions that can accumulate through those interactions, and ultimately drift away from the correct answer.
\newline
\newline
\noindent \textbf{Image Key-Value Cache Dropping:} We verify the above through causal intervention based on image token dropping. Specifically, we take 300 random samples for each dataset, and, at different generation steps, completely remove the image tokens (i.e., their Key/Value cache) from all layers, and continue generation. We then measure the accuracy for each drop step. We consider drop steps at \{20, 50, 100, 200 and 300\}. The results are shown in Figure \ref{fig:dop_kv_cache}, where the x-axis represents the drop step and the y-axis represents the accuracy. Dropping image tokens at early generation steps leads to a clear performance degradation. This is because visual information is removed during the critical front-loading phase, when it is encoded and absorbed into the text-token representations. When image tokens are dropped later in the generation process, the effect becomes much less pronounced. After approximately 200 generation steps, removing the image tokens has almost no effect on accuracy. Intervention results for other datasets are presented in Section \ref{supp:kv_cache_drop_additional} of the Supp. material. We also provide computational savings (in GFLOPs) in Section \ref{supp:kv_cache_drop_flops} of the Supp. material.

\Finding{3}{Long CoTs rely on early visual encoding}{
During CoT generation, attention to image tokens is strongest at the beginning and gradually declines, while attention to previously generated tokens increases. This suggests that later steps depend on visual information encoded early and absorbed into the text token representations, relying more on prior textual context for reasoning rather than repeatedly attending to the image.
}
\vspace{-0.5cm}

\section{Analysis of Multimodal Chain-of-Thoughts}
\textbf{LLM-as-a-Judge:}
To evaluate the quality of the generated multimodal rationales, we employ an external LLM as a judge, and later validate its agreement with human evaluation. Given a question $q$, a ground-truth answer $x$ and and a rationale $r$ of length $Y$ produced by the LVLM with CoT prompting, we measure two complementary metrics: \textbf{(1) Rationale Sufficiency}, where the judge receives $(q, r)$ and predicts an answer $\hat{x}_r$ based solely on the textual content of $r$, without access to the image, external knowledge or the ground-truth answer $x$. The idea is that if a judge can answer the question solely by reading the reasoning chain $r$, then $r$ is highly informative and contains sufficient information. The rationale accuracy is computed by comparing $\hat{x}_r$ to $x$. We then compare this rationale accuracy with that of the LVLMs. \textbf{(2) Rationale Dynamics}, a fine-grained analysis of (1), where the rationale $r$ is decomposed into its components, represented by an ordered sequence of sentences $\{r_1,\dots,r_S\}$ where $S$ represents the total number of sentences in $r$. Each sentence $r_s$ is evaluated independently to determine whether its information is sufficient enough to supports a positive answer, a negative answer, or an uncertain answer with respect to $q$. From these sentence-level signals, we derive aggregate statistics including the earliest decision point, shifts or contradictions across reasoning steps, patterns of intermediate uncertainty, and the position of final commitment to the answer within the rationale. All prompts are provided in Section \ref{supp:llm-judge-prompts} of the Supp. material. 

\begin{figure}
    \centering
    \includegraphics[width=1\linewidth]{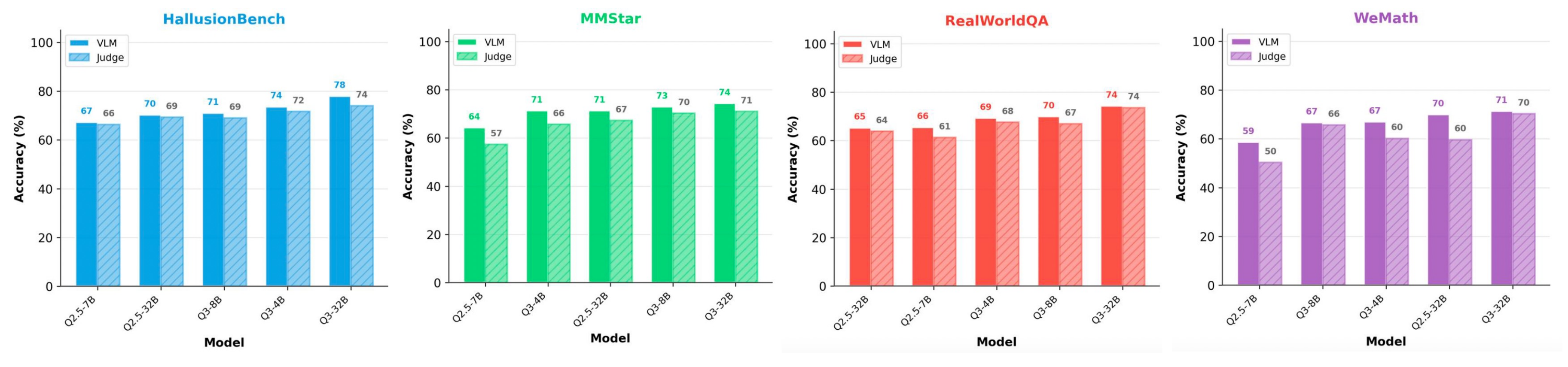}
    \caption{LVLM vs Judge accuracy across benchmarks and model scales. Results are computed on 300 random samples per benchmark with $Y=300$ tokens.}
    \label{fig:vlm_judge_acc}
\end{figure}

\noindent \textbf{Rationale Sufficiency:} The results are presented in Figure~\ref{fig:vlm_judge_acc}. Across most benchmarks and model scales, judge accuracy closely follows overall LVLM performance, indicating that the rationales capture a meaningful decision-relevant signal. On WeMath, there exists some models with a small gap between the LVLM performance and the judge accuracy (e.g., Qwen-2.5-VL 7B and 32B), where rationale-level accuracy lags behind LVLM accuracy, indicating that intermediate reasoning steps are not fully expressed. In general, we find that higher quality models (either in terms of size or generation) often narrow the gap, indicating better explainability via more informative rationales. 
\newline
\newline
\noindent \textbf{Human Evaluation:} We also perform human evaluation to assess the reliability of the LLM judge. The results are presented in Section \ref{supp:human_validation} of the Supp. material and indicate high consistency and strong agreement with the judge, supporting the judge use as a reliable proxy for evaluating rationale sufficiency.

\begin{figure}
    \centering
    \includegraphics[width=1\linewidth]{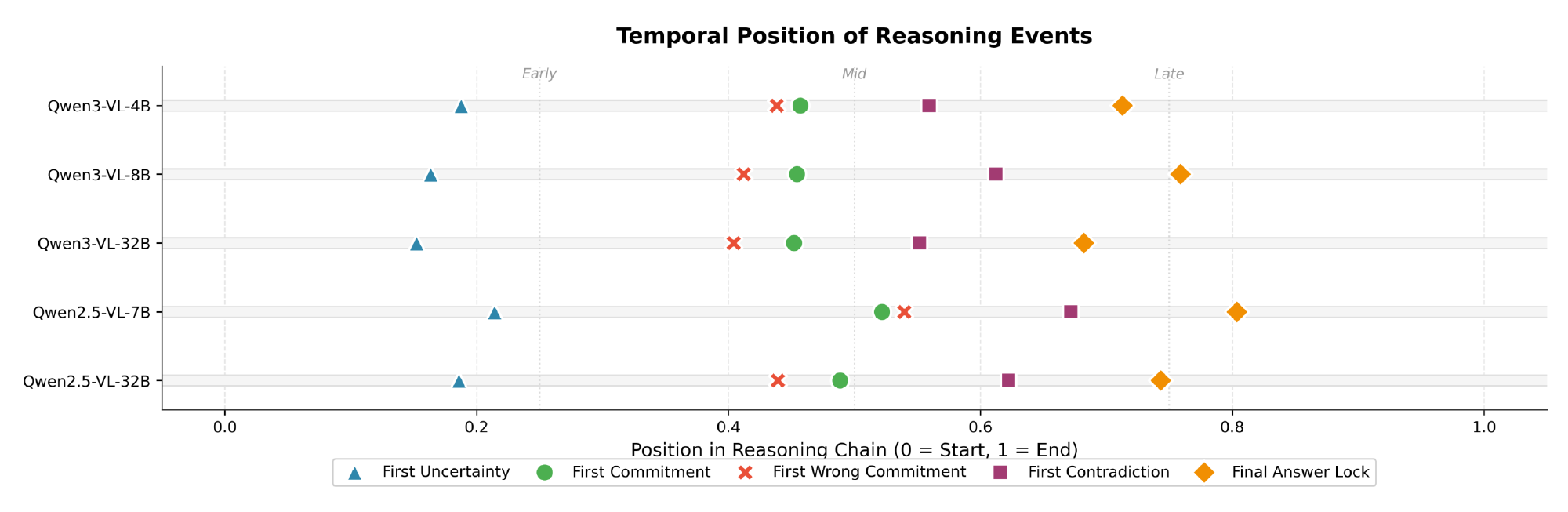}
    \caption{Analysis of LLM-Judge evaluated rationale dynamics per different model on HallusionBench.}
    \label{fig:reasoning_sentence}
\end{figure}

\par\vspace{1\baselineskip}
\noindent \textbf{Rationale Dynamics:} We focus on HallusionBench as a representative benchmark combining perception, reasoning and hallucination. However, similar trends are observed on other benchmarks. Figure \ref{fig:reasoning_sentence} shows the temporal position of key reasoning events on HallusionBench, measured as normalized position within the rationale (0 = start, 1 = end). According to the judge's predictions, uncertainty typically appears early in the chain ({\color{blue} $\blacktriangle$}). That is, there is uncertain information at that point in the chain to derive an answer. Explicit support for the correct answer then starts to appear midway in the chain ({\color{green} $\bullet$}). That is, the chain becomes more informative. This happens in around 75\% of the samples. In the other 25\% of instances, there is still no support for the correct answer ({\color{red} $\times$}). However, for those instances, the information content improves at around 0.6 of the chain ({\color{purple} $\blacksquare$}) giving support for the correct answer. Shortly after, at around 0.75 of the chain ({\color{orange} $\blacklozenge$}), the information is sufficient and informative enough, such that an answer can be derived without being changed in later steps. The observed temporal structure suggests that generated rationales tend to organize reasoning (through information sufficiency) progressively, rather than presenting a fully formed conclusion from the outset. 

\section{Conclusion}
We presented the first broad study of zero-shot test-time scaling for LVLMs. We find that standard LLM test-time scaling methods transfer well to LVLMs and are often \emph{most} effective for small instruction-following models, yielding large gains. However, extra compute can hurt on perception-focused tasks, where long generations cause models to overthink and loose focus. We also show that visual information is encoded early, and image tokens contribute less as generation progresses. Overall, zero-shot test-time scaling is a strong accuracy--compute lever for LVLMs, particularity for small models, but should be applied carefully.

\section*{Acknowledgements}
{\footnotesize
Fawaz Sammani is funded by the Fonds Wetenschappelijk Onderzoek (FWO) (PhD) fellowship strategic basic research 1SH7W24N). T. Chamiti and N. Deligiannis acknowledge the ”Onderzoeksprogramma Artificiele Intelligentie (AI) Vlaanderen” programme and the ERC Consolidator Grant IONIAN (No. 101171240, DOI: 10.3030/101171240). Funded by the European Union. Views and opinions expressed are however those of the author(s) only and do not necessarily reflect those of the European Union or the European Research Council Executive Agency. Neither the European Union nor the granting authority can be held responsible for them.
}

\bibliographystyle{splncs04}
\bibliography{main}

\title{On Test-Time Scaling for Vision-Language Models (Supplementary Material)}
\titlerunning{Supplementary Material}

\author{}
\institute{}
\maketitle

\renewcommand{\thesection}{S\arabic{section}}
\renewcommand{\thefigure}{S\arabic{figure}}
\renewcommand{\thetable}{S\arabic{table}}

\renewcommand{\theHsection}{supp.section.\arabic{section}}
\renewcommand{\theHsubsection}{supp.subsection.\arabic{section}.\arabic{subsection}}
\renewcommand{\theHfigure}{supp.figure.\arabic{figure}}
\renewcommand{\theHtable}{supp.table.\arabic{table}}

\setcounter{section}{0}
\setcounter{figure}{0}
\setcounter{table}{0}

\section{InternVL-3.5 Results}
\label{supp:intnervl}
Tabular results on the InternVL-3.5 series are presented in Table \ref{tab:internvl_tts_results}.  

\begin{table*}[ph]
\centering
\caption{
Test-time scaling results for InternVL models.
\colorbox{red!25}{Degradation} indicates a score lower than the baseline, while
\colorbox{green!25}{Improvement} indicates a score higher than the baseline.
}
\setlength{\tabcolsep}{4pt}
\renewcommand{\arraystretch}{1.15}
\footnotesize

\scalebox{0.89}{
\begin{tabular}{c l *{5}{c}}
\toprule
\multirow{2}{*}{\rotatebox[origin=c]{90}{Model}}
& \multirow{2}{*}{Method}
& \multicolumn{5}{c}{Dataset} \\
\cmidrule(lr){3-7}
&
& MMStar
& RealWorldQA
& HallusionBench
& LogicVista
& WeMath \\
\midrule

\multirow{7}{*}{\rotatebox[origin=c]{90}{\textbf{InternVL-3.5-2B}}}
& Baseline & 59.40 & 60.65 & 62.04 & 31.25 & 48.51 \\
& CoT & \cellcolor{green!67}63.86 & \cellcolor{red!21}59.22 & \cellcolor{green!85}69.51 & \cellcolor{green!85}40.62 & \cellcolor{green!85}57.70 \\
& S-CoT & \cellcolor{green!57}63.17 & \cellcolor{red!29}58.69 & \cellcolor{green!85}69.72 & \cellcolor{green!85}45.09 & \cellcolor{green!85}56.78 \\
& PaS & \cellcolor{green!64}63.65 & \cellcolor{green!18}61.83 & \cellcolor{green!85}68.14 & \cellcolor{green!85}43.30 & \cellcolor{green!85}58.16 \\
& Self-Refinement & \cellcolor{green!57}63.18 & \cellcolor{red!33}58.43 & \cellcolor{green!85}69.09 & \cellcolor{green!85}39.51 & \cellcolor{green!85}57.82 \\
& Self-Consistency & \cellcolor{green!85}65.78 & \cellcolor{green!20}61.96 & \cellcolor{green!85}69.19 & \cellcolor{green!85}45.54 & \cellcolor{green!85}64.54 \\
& Self-Aggregation & \cellcolor{green!62}63.53 & 60.65 & \cellcolor{green!85}68.87 & \cellcolor{green!85}42.63 & \cellcolor{green!85}60.80 \\

\midrule

\multirow{7}{*}{\rotatebox[origin=c]{90}{\textbf{InternVL-3.5-4B}}}
& Baseline & 66.34 & 66.01 & 67.93 & 37.72 & 56.21 \\
& CoT & \cellcolor{green!15}66.48 & \cellcolor{red!85}51.63 & \cellcolor{green!43}70.77 & \cellcolor{green!85}50.45 & \cellcolor{green!85}70.23 \\
& S-CoT & \cellcolor{green!18}67.51 & \cellcolor{red!85}52.03 & \cellcolor{green!57}71.71 & \cellcolor{green!85}51.12 & \cellcolor{green!85}69.94 \\
& PaS & \cellcolor{green!15}66.59 & \cellcolor{red!85}49.28 & \cellcolor{green!17}69.09 & \cellcolor{green!85}50.89 & \cellcolor{green!85}71.15 \\
& Self-Refinement & \cellcolor{green!15}66.48 & \cellcolor{red!85}50.59 & \cellcolor{green!54}71.50 & \cellcolor{green!85}52.90 & \cellcolor{green!85}69.77 \\
& Self-Consistency & \cellcolor{green!36}68.72 & \cellcolor{red!85}50.20 & \cellcolor{green!54}71.50 & \cellcolor{green!85}53.35 & \cellcolor{green!85}74.54 \\
& Self-Aggregation & \cellcolor{green!30}68.33 & \cellcolor{red!85}44.31 & \cellcolor{green!30}69.93 & \cellcolor{green!85}52.68 & \cellcolor{green!85}71.55 \\

\midrule

\multirow{7}{*}{\rotatebox[origin=c]{90}{\textbf{InternVL-3.5-8B}}}
& Baseline & 67.01 & 64.18 & 70.77 & 38.39 & 61.32 \\
& CoT & \cellcolor{green!15}67.56 & \cellcolor{red!85}54.64 & \cellcolor{green!15}71.71 & \cellcolor{green!85}48.88 & \cellcolor{green!85}69.54 \\
& S-CoT & \cellcolor{red!15}66.48 & \cellcolor{red!63}60.00 & \cellcolor{green!15}71.71 & \cellcolor{green!85}47.77 & \cellcolor{green!85}70.46 \\
& PaS & \cellcolor{green!15}67.71 & \cellcolor{red!85}53.46 & \cellcolor{green!38}73.29 & \cellcolor{green!85}48.66 & \cellcolor{green!85}71.21 \\
& Self-Refinement & \cellcolor{red!15}66.76 & \cellcolor{red!85}55.29 & \cellcolor{green!43}73.61 & \cellcolor{green!85}49.78 & \cellcolor{green!85}70.34 \\
& Self-Consistency & \cellcolor{green!45}69.99 & \cellcolor{red!85}56.73 & \cellcolor{green!15}71.61 & \cellcolor{green!85}52.90 & \cellcolor{green!85}71.09 \\
& Self-Aggregation & \cellcolor{green!19}68.27 & \cellcolor{red!85}49.54 & \cellcolor{green!27}72.56 & \cellcolor{green!85}54.69 & \cellcolor{green!85}71.49 \\

\midrule

\multirow{7}{*}{\rotatebox[origin=c]{90}{\textbf{InternVL-3.5-38B}}}
& Baseline & 73.18 & 73.73 & 72.66 & 44.64 & 65.80 \\
& CoT & \cellcolor{red!15}72.45 & \cellcolor{red!26}72.03 & \cellcolor{green!24}74.24 & \cellcolor{green!85}56.03 & \cellcolor{green!85}73.97 \\
& S-CoT & \cellcolor{green!15}73.73 & \cellcolor{red!30}71.76 & \cellcolor{green!27}74.45 & \cellcolor{green!85}57.14 & \cellcolor{green!51}69.20 \\
& PaS & \cellcolor{green!15}74.08 & \cellcolor{red!15}73.59 & \cellcolor{green!24}74.24 & \cellcolor{green!85}57.37 & \cellcolor{green!85}72.41 \\
& Self-Refinement & \cellcolor{red!18}72.00 & \cellcolor{red!77}68.63 & \cellcolor{green!15}73.61 & \cellcolor{green!85}57.59 & \cellcolor{green!85}74.08 \\
& Self-Consistency & \cellcolor{green!29}75.13 & \cellcolor{red!18}72.55 & \cellcolor{green!24}74.24 & \cellcolor{green!85}62.95 & \cellcolor{green!85}77.13 \\
& Self-Aggregation & \cellcolor{red!16}72.09 & \cellcolor{red!43}70.85 & \cellcolor{red!17}71.50 & \cellcolor{green!85}59.38 & \cellcolor{green!85}75.11 \\

\bottomrule
\end{tabular}
}
\label{tab:internvl_tts_results}
\end{table*}

\section{Molmo2 Results}
\label{supp:molmo2}

We present results on the Molmo2 model in the \{4B, 8B\} scale in Table \ref{tab:molmo_tts_results}. Across the Molmo2 series, we find that the overall trend remains generally consistent with that of the Qwen series presented in the main manuscript.
\begin{table*}[h]
\centering
\caption{Test-Time scaling results on the Molmo series, with \colorbox{red!32}{degradation} meaning a lower score than baseline, and \colorbox{green!32}{improvement} meaning a higher score than baseline.}
\setlength{\tabcolsep}{4pt}
\renewcommand{\arraystretch}{1.15}
\footnotesize
\scalebox{0.98}{
\begin{tabular}{c l *{5}{c}}
\toprule
\multirow{2}{*}{\rotatebox[origin=c]{90}{Model}} & \multirow{2}{*}{Method} & \multicolumn{5}{c}{Dataset} \\
\cmidrule(lr){3-7}
 &  & MMStar & RealWorldQA & HallusionBench & WeMath & LogicVista \\
\midrule
\multirow{9}{*}{\rotatebox[origin=c]{90}{\textbf{Molmo2-4B}}}
& Baseline & 62.45 & 73.33 & 68.56 & 45.52 & 35.49 \\
& CoT
& \cellcolor{green!22}{65.63}
& \cellcolor{red!41}{64.44}
& \cellcolor{red!20}{66.88}
& \cellcolor{green!35}{52.53}
& \cellcolor{green!34}{42.19} \\
& S-CoT
& \cellcolor{red!20}{62.38}
& \cellcolor{red!72}{55.03}
& \cellcolor{green!20}{70.03}
& \cellcolor{green!26}{49.89}
& \cellcolor{green!32}{41.74} \\
& Self-Consistency 
& \cellcolor{green!22}{65.37}
& \cellcolor{red!29}{68.10}
& \cellcolor{green!20}{70.24}
& \cellcolor{green!39}{53.85}
& \cellcolor{green!45}{45.54} \\
& Self-Refinement 
& \cellcolor{red!20}{61.74}
& \cellcolor{red!45}{63.27}
& \cellcolor{red!21}{65.83}
& \cellcolor{green!36}{52.70}
& \cellcolor{green!32}{41.74} \\
& Self-Aggregation 
& \cellcolor{red!20}{61.07}
& \cellcolor{red!54}{60.65}
& \cellcolor{red!20}{66.46}
& \cellcolor{green!53}{57.93}
& \cellcolor{green!52}{47.77} \\
& Describe-Answer 
& \cellcolor{red!20}{60.23}
& \cellcolor{red!20}{73.07}
& \cellcolor{red!20}{67.40}
& \cellcolor{green!20}{46.03}
& \cellcolor{red!20}{35.04} \\
& CCoT 
& \cellcolor{green!20}{62.75}
& \cellcolor{red!33}{66.80}
& \cellcolor{red!26}{64.25}
& \cellcolor{red!20}{43.39}
& \cellcolor{red!20}{34.15} \\
& Prompt Repetition 
& \cellcolor{green!20}{63.07}
& \cellcolor{red!20}{73.20}
& \cellcolor{red!20}{66.77}
& \cellcolor{red!28}{40.52}
& \cellcolor{red!20}{35.27} \\

\midrule
\multirow{9}{*}{\rotatebox[origin=c]{90}{\textbf{Molmo2-8B}}}
& Baseline & 64.00 & 75.29 & 68.24 & 53.79 & 35.71 \\
& CoT 
& \cellcolor{red!20}{62.96}
& \cellcolor{red!24}{71.50}
& \cellcolor{red!20}{66.98}
& \cellcolor{green!23}{57.24}
& \cellcolor{green!32}{41.96} \\
& S-CoT
& \cellcolor{red!20}{61.71}
& \cellcolor{red!40}{66.80}
& \cellcolor{red!20}{67.72}
& \cellcolor{green!20}{56.38}
& \cellcolor{green!38}{43.53} \\
& Self-Consistency 
& \cellcolor{green!23}{67.41}
& \cellcolor{green!20}{76.08}
& \cellcolor{green!20}{68.87}
& \cellcolor{green!32}{59.94}
& \cellcolor{green!36}{43.08} \\
& Self-Refinement
& \cellcolor{red!20}{63.06}
& \cellcolor{red!39}{66.93}
& \cellcolor{red!20}{67.72}
& \cellcolor{green!23}{57.07}
& \cellcolor{green!41}{44.42} \\
& Self-Aggregation 
& \cellcolor{red!20}{62.79}
& \cellcolor{red!33}{69.02}
& \cellcolor{green!20}{68.56}
& \cellcolor{green!26}{57.99}
& \cellcolor{green!46}{46.21} \\
& Describe-Answer 
& \cellcolor{green!20}{64.26}
& \cellcolor{red!20}{74.12}
& \cellcolor{green!20}{70.35}
& \cellcolor{red!20}{52.93}
& \cellcolor{red!20}{35.49} \\
& CCoT 
& \cellcolor{red!20}{63.23}
& \cellcolor{red!23}{71.90}
& \cellcolor{red!23}{64.98}
& \cellcolor{red!23}{50.57}
& \cellcolor{red!20}{35.49} \\
& Prompt Repetition
& \cellcolor{green!20}{64.32}
& \cellcolor{green!20}{76.34}
& \cellcolor{green!20}{68.87}
& \cellcolor{red!42}{44.54}
& \cellcolor{red!20}{34.60} \\
\bottomrule
\end{tabular}
}
\label{tab:molmo_tts_results}
\end{table*}

\section{Additional Benchmarks and Perception Benchmarks Degradation}

\subsection{A-OKVQA Results}
\label{supp:aokvqa}
We report results on the A-OKVQA benchmark, a perception-only benchmark, in Table \ref{tab:aokvqa_results}. These experiments further support the claim made in the main manuscript that test-time scaling may degrade performance, or at least fail to improve it, on benchmarks that require little or no reasoning (in general, benchmarks that require additional computation). As shown in Table~\ref{tab}, most test-time scaling methods perform below the baseline (indicated in red) while the remaining methods show no improvement over the baseline. Only a few methods in the Qwen3-VL-4B series achieve slight improvements over the baseline.

\begin{table*}[h]
\centering
\caption{A-OKVQA accuracy results. Cell color intensity is proportional to the absolute change from the baseline within each model column, with a minimum visible intensity for degradations.}
\resizebox{\textwidth}{!}{%
\begin{tabular}{lccccccccc}
\toprule
Method & \rotatebox{90}{Qwen3-VL-2B-Instruct} & \rotatebox{90}{Qwen3-VL-4B-Instruct} & \rotatebox{90}{Qwen3-VL-8B-Instruct} & \rotatebox{90}{Qwen3-VL-32B-Instruct} & \rotatebox{90}{Qwen2.5-VL-7B-Instruct} & \rotatebox{90}{Qwen2.5-VL-32B-Instruct} & \rotatebox{90}{Qwen2.5-VL-72B-Instruct} & \rotatebox{90}{Molmo2-4B} & \rotatebox{90}{Molmo2-8B} \\
\midrule
Baseline & 81.92 & 85.15 & 87.16 & 89.17 & 87.77 & 86.20 & 90.92 & 88.21 & 89.34 \\
CoT & \cellcolor{red!27}81.40 & \cellcolor{green!22}86.11 & \cellcolor{red!31}86.46 & \cellcolor{red!42}87.95 & \cellcolor{red!58}85.85 & \cellcolor{red!23}85.85 & \cellcolor{red!78}88.12 & \cellcolor{red!31}87.51 & \cellcolor{red!48}87.86 \\
S-CoT & \cellcolor{red!90}78.60 & \cellcolor{green!33}86.64 & \cellcolor{red!29}86.55 & \cellcolor{red!21}88.91 & \cellcolor{red!40}86.64 & \cellcolor{red!44}84.89 & \cellcolor{red!72}88.38 & \cellcolor{red!72}85.68 & \cellcolor{red!70}86.90 \\
PaS & \cellcolor{red!54}80.17 & \cellcolor{green!14}85.76 & \cellcolor{green!0}87.16 & \cellcolor{red!27}88.65 & \cellcolor{red!76}85.07 & \cellcolor{red!42}84.98 & \textemdash & \textemdash & \textemdash \\
Self-Consistency & \cellcolor{green!39}83.67 & \cellcolor{green!53}87.51 & \cellcolor{green!4}87.34 & \cellcolor{green!10}89.61 & \cellcolor{red!46}86.38 & \cellcolor{green!0}86.20 & \cellcolor{red!43}89.69 & \cellcolor{green!29}89.52 & \cellcolor{green!12}89.87 \\
Self-Aggregation & \cellcolor{green!18}82.71 & \cellcolor{green!41}86.99 & \cellcolor{red!35}86.29 & \cellcolor{green!12}89.69 & \cellcolor{red!66}85.50 & \cellcolor{red!17}86.11 & \cellcolor{red!56}89.08 & \cellcolor{red!35}87.34 & \cellcolor{red!37}88.38 \\
Self-Refinement & \cellcolor{red!19}81.75 & \cellcolor{green!4}85.33 & \cellcolor{red!48}85.68 & \cellcolor{red!60}87.16 & \cellcolor{red!66}85.50 & \cellcolor{red!33}85.41 & \cellcolor{red!95}87.34 & \cellcolor{red!72}85.68 & \cellcolor{red!29}88.73 \\
Describe-Answer & \cellcolor{red!29}81.31 & \cellcolor{green!12}85.68 & \cellcolor{red!17}87.07 & \cellcolor{red!17}89.08 & \cellcolor{red!39}86.72 & \cellcolor{green!6}86.46 & \cellcolor{red!37}89.96 & \cellcolor{green!4}88.38 & \cellcolor{green!16}90.04 \\
CCoT & \cellcolor{red!58}80.00 & \cellcolor{red!17}85.07 & \cellcolor{red!42}85.94 & \cellcolor{red!27}88.65 & \cellcolor{red!100}83.32 & \cellcolor{red!29}85.59 & \cellcolor{red!60}88.91 & \cellcolor{red!41}87.07 & \cellcolor{red!31}88.65 \\
Prompt Repetition & \cellcolor{green!2}82.01 & \cellcolor{green!20}86.03 & \cellcolor{green!20}88.03 & \cellcolor{green!2}89.26 & \cellcolor{red!50}86.20 & \cellcolor{green!4}86.38 & \cellcolor{green!2}91.00 & \cellcolor{red!19}88.03 & \cellcolor{green!8}89.69 \\
\bottomrule
\end{tabular}%
}
\label{tab:aokvqa_results}
\end{table*}

\subsection{MMStar Categorical Scores}
\label{supp:mmstar_categorical_perception_degrade}

We report the results for the different categories of the MMStar benchmark. MMStar comprises six categories. The perception-focused categories are coarse perception, fine-grained perception, and science \& technology; these tasks typically require little or no reasoning. The remaining three categories assess reasoning ability: instance reasoning, logical reasoning, and mathematics. These tasks require more explicit reasoning and deliberation. The results are presented in Table \ref{tab:mmstar_categ}. Overall, they are consistent with the findings in the main manuscript, showing that test-time scaling often degrades performance on perception-oriented benchmarks. 

\newpage
{\footnotesize
\begin{longtable}{c l r r r r r r r}
\caption{MMStar results (values in \% , rounded to two decimals). Results (higher is better) with cell colors indicating change relative to the model’s baseline on each dataset: \protect\colorbox{red!25}{degradation} indicates a lower score than baseline, and \protect\colorbox{green!25}{improvement} indicates a higher score than baseline (darker shades indicate larger changes).}
\label{tab:mmstar_categ}\\
\toprule
\textbf{Model} & \textbf{Method} &
\rotatebox{90}{\makecell{Coarse \\ Perception}} &
\rotatebox{90}{\makecell{Finegrained \\ Perception}} &
\rotatebox{90}{\makecell{Instance \\ Reasoning}} &
\rotatebox{90}{\makecell{Logical \\ Reasoning}} &
\rotatebox{90}{\makecell{Math}} &
\rotatebox{90}{\makecell{Science}} &
\rotatebox{90}{\makecell{Overall}} \\
\midrule
\endfirsthead

\toprule
\textbf{Model} & \textbf{Method} &
\rotatebox{90}{\makecell{Coarse \\ Perception}} &
\rotatebox{90}{\makecell{Finegrained \\ Perception}} &
\rotatebox{90}{\makecell{Instance \\ Reasoning}} &
\rotatebox{90}{\makecell{Logical \\ Reasoning}} &
\rotatebox{90}{\makecell{Math}} &
\rotatebox{90}{\makecell{Science}} &
\rotatebox{90}{\makecell{Overall}} \\
\midrule
\endhead

\midrule
\multicolumn{9}{r}{\small\itshape Continued on next page}\\
\midrule
\endfoot

\bottomrule
\endlastfoot

\multirow{9}{*}{\rotatebox[origin=c]{90}{\textbf{Qwen2.5-VL-7B}}} & Baseline & 71.57 & 59.64 & 71.71 & 62.46 & 66.16 & 43.90 & 62.57 \\
& CoT & \cellcolor{red!33}70.36 & \cellcolor{red!66}53.28 & \cellcolor{green!25}71.76 & \cellcolor{green!59}67.64 & \cellcolor{green!64}72.23 & \cellcolor{green!35}45.49 & \cellcolor{green!31}63.46 \\
& S-CoT & \cellcolor{green!28}72.04 & \cellcolor{red!63}53.77 & \cellcolor{green!29}72.37 & \cellcolor{green!62}68.19 & \cellcolor{green!82}74.85 & \cellcolor{red!28}43.46 & \cellcolor{green!35}64.11 \\
& Self-Consistency & \cellcolor{red!37}69.68 & \cellcolor{red!31}58.74 & \cellcolor{green!39}73.83 & \cellcolor{green!78}70.64 & \cellcolor{green!81}74.82 & \cellcolor{green!26}44.05 & \cellcolor{green!43}65.29 \\
& Describe-Answer & \cellcolor{red!25}71.53 & \cellcolor{red!41}57.23 & \cellcolor{red!34}70.26 & \cellcolor{red!45}59.32 & \cellcolor{red!47}62.85 & \cellcolor{red!42}41.30 & \cellcolor{red!39}60.41 \\
& Self-Aggregation & \cellcolor{red!37}69.71 & \cellcolor{red!80}51.17 & \cellcolor{red!45}68.70 & \cellcolor{green!71}69.59 & \cellcolor{green!52}70.37 & \cellcolor{red!33}42.71 & \cellcolor{red!28}62.04 \\
& Self-Refinement & \cellcolor{red!30}70.82 & \cellcolor{red!65}53.54 & \cellcolor{green!32}72.85 & \cellcolor{red!41}59.95 & \cellcolor{green!55}70.78 & \cellcolor{red!50}40.06 & \cellcolor{red!33}61.33 \\
& CCoT & \cellcolor{red!36}69.90 & \cellcolor{red!55}55.09 & \cellcolor{red!50}67.80 & \cellcolor{red!31}61.60 & \cellcolor{red!29}65.51 & \cellcolor{red!27}43.59 & \cellcolor{red!38}60.58 \\
& Prompt Repetition & \cellcolor{green!30}72.28 & \cellcolor{red!59}54.40 & \cellcolor{red!64}65.75 & \cellcolor{red!34}61.06 & \cellcolor{red!46}62.95 & \cellcolor{green!38}45.85 & \cellcolor{red!39}60.38 \\
\midrule
\multirow{9}{*}{\rotatebox[origin=c]{90}{\textbf{Qwen2.5-VL-32B}}} & Baseline & 75.77 & 56.41 & 73.46 & 70.20 & 77.02 & 51.27 & 67.35 \\
& CoT & \cellcolor{red!69}69.03 & \cellcolor{green!26}56.60 & \cellcolor{red!26}73.30 & \cellcolor{green!39}72.29 & \cellcolor{green!51}80.98 & \cellcolor{red!39}49.04 & \cellcolor{red!28}66.87 \\
& S-CoT & \cellcolor{red!57}70.86 & \cellcolor{green!40}58.79 & \cellcolor{green!31}74.46 & \cellcolor{green!43}73.04 & \cellcolor{green!59}82.25 & \cellcolor{red!28}50.80 & \cellcolor{green!32}68.37 \\
& Self-Consistency & \cellcolor{red!45}72.75 & \cellcolor{green!33}57.62 & \cellcolor{green!43}76.18 & \cellcolor{green!46}73.48 & \cellcolor{green!60}82.43 & \cellcolor{green!26}51.37 & \cellcolor{green!36}68.97 \\
& Describe-Answer & \cellcolor{green!29}76.37 & \cellcolor{green!26}56.62 & \cellcolor{red!38}71.46 & \cellcolor{red!37}68.36 & \cellcolor{red!58}71.98 & \cellcolor{green!37}53.10 & \cellcolor{red!32}66.32 \\
& Self-Aggregation & \cellcolor{red!55}71.20 & \cellcolor{red!33}55.14 & \cellcolor{green!31}74.33 & \cellcolor{green!39}72.40 & \cellcolor{green!58}82.02 & \cellcolor{red!47}47.85 & \cellcolor{red!26}67.16 \\
& Self-Refinement & \cellcolor{red!77}67.83 & \cellcolor{red!30}55.66 & \cellcolor{green!36}75.09 & \cellcolor{red!26}70.12 & \cellcolor{green!45}80.02 & \cellcolor{red!66}45.02 & \cellcolor{red!36}65.62 \\
& CCoT & \cellcolor{red!43}72.94 & \cellcolor{red!53}52.15 & \cellcolor{red!30}72.72 & \cellcolor{green!32}71.30 & \cellcolor{red!47}73.70 & \cellcolor{red!29}50.68 & \cellcolor{red!37}65.58 \\
& Prompt Repetition & \cellcolor{green!34}77.19 & \cellcolor{green!25}56.44 & \cellcolor{green!28}73.91 & \cellcolor{red!39}68.11 & \cellcolor{red!30}76.26 & \cellcolor{red!37}49.35 & \cellcolor{red!28}66.88 \\
\midrule
\multirow{9}{*}{\rotatebox[origin=c]{90}{\textbf{Qwen2.5-VL-72B}}} & Baseline & 76.14 & 59.75 & 73.21 & 75.30 & 76.49 & 51.01 & 68.65 \\
& CoT & \cellcolor{red!40}73.89 & \cellcolor{red!34}58.32 & \cellcolor{red!26}73.01 & \cellcolor{green!41}77.75 & \cellcolor{green!50}80.33 & \cellcolor{red!43}48.20 & \cellcolor{red!25}68.58 \\
& S-CoT & \cellcolor{red!50}72.30 & \cellcolor{red!44}56.86 & \cellcolor{green!25}73.29 & \cellcolor{green!41}77.69 & \cellcolor{green!50}80.31 & \cellcolor{green!33}52.18 & \cellcolor{green!26}68.77 \\
& Self-Consistency & \cellcolor{red!26}75.92 & \cellcolor{red!42}57.12 & \cellcolor{green!33}74.43 & \cellcolor{green!70}82.18 & \cellcolor{green!43}79.31 & \cellcolor{green!59}56.20 & \cellcolor{green!39}70.86 \\
& Describe-Answer & \cellcolor{red!52}72.02 & \cellcolor{red!33}58.48 & \cellcolor{green!26}73.41 & \cellcolor{red!57}70.42 & \cellcolor{red!34}75.04 & \cellcolor{green!27}51.30 & \cellcolor{red!37}66.78 \\
& Self-Aggregation & \cellcolor{red!28}75.73 & \cellcolor{red!32}58.64 & \cellcolor{green!47}76.54 & \cellcolor{green!43}78.10 & \cellcolor{green!47}79.81 & \cellcolor{green!43}53.82 & \cellcolor{green!37}70.44 \\
& Self-Refinement & \cellcolor{red!35}74.64 & \cellcolor{green!38}61.77 & \cellcolor{red!25}73.18 & \cellcolor{green!39}77.44 & \cellcolor{green!57}81.43 & \cellcolor{red!65}44.89 & \cellcolor{green!27}68.89 \\
& CCoT & \cellcolor{red!33}74.93 & \cellcolor{green!38}61.78 & \cellcolor{red!26}73.12 & \cellcolor{red!49}71.61 & \cellcolor{green!30}77.32 & \cellcolor{red!37}49.18 & \cellcolor{red!29}67.99 \\
& Prompt Repetition & \cellcolor{green!38}78.16 & \cellcolor{green!33}60.98 & \cellcolor{red!30}72.50 & \cellcolor{red!35}73.78 & \cellcolor{green!28}76.98 & \cellcolor{green!45}54.01 & \cellcolor{green!30}69.40 \\
\midrule
\multirow{9}{*}{\rotatebox[origin=c]{90}{\textbf{Qwen3-VL-32B}}} & Baseline & 77.01 & 68.59 & 78.40 & 75.38 & 71.42 & 65.14 & 72.66 \\
& CoT & \cellcolor{red!32}75.86 & \cellcolor{red!37}66.72 & \cellcolor{green!35}80.01 & \cellcolor{green!57}80.25 & \cellcolor{green!90}86.93 & \cellcolor{red!62}59.52 & \cellcolor{green!39}74.88 \\
& S-CoT & \cellcolor{red!43}74.18 & \cellcolor{red!33}67.38 & \cellcolor{red!38}76.39 & \cellcolor{green!58}80.53 & \cellcolor{green!90}84.95 & \cellcolor{red!29}64.60 & \cellcolor{green!38}74.67 \\
& Self-Consistency & \cellcolor{green!31}77.88 & \cellcolor{green!26}68.76 & \cellcolor{green!39}80.56 & \cellcolor{green!72}82.60 & \cellcolor{green!90}86.13 & \cellcolor{red!40}62.83 & \cellcolor{green!50}76.46 \\
& Describe-Answer & \cellcolor{green!28}77.44 & \cellcolor{red!37}66.77 & \cellcolor{red!28}77.90 & \cellcolor{red!40}73.01 & \cellcolor{red!90}61.05 & \cellcolor{red!26}64.99 & \cellcolor{red!41}70.20 \\
& Self-Aggregation & \cellcolor{red!31}76.16 & \cellcolor{red!31}67.63 & \cellcolor{red!33}77.19 & \cellcolor{green!90}85.52 & \cellcolor{green!90}86.54 & \cellcolor{red!45}62.04 & \cellcolor{green!46}75.85 \\
& Self-Refinement & \cellcolor{red!55}72.34 & \cellcolor{red!53}64.26 & \cellcolor{green!26}78.59 & \cellcolor{green!54}79.79 & \cellcolor{green!90}86.97 & \cellcolor{red!76}57.31 & \cellcolor{green!29}73.21 \\
& CCoT & \cellcolor{red!26}76.81 & \cellcolor{red!27}68.34 & \cellcolor{green!45}81.52 & \cellcolor{red!32}74.31 & \cellcolor{green!44}74.41 & \cellcolor{red!67}58.75 & \cellcolor{red!27}72.36 \\
& Prompt Repetition & 77.01 & \cellcolor{green!26}68.68 & \cellcolor{green!28}78.94 & \cellcolor{red!61}69.90 & \cellcolor{green!27}71.75 & \cellcolor{green!32}66.17 & \cellcolor{red!29}72.07 \\
\midrule
\multirow{9}{*}{\rotatebox[origin=c]{90}{\textbf{Qwen3-VL-2B}}} & Baseline & 72.75 & 54.54 & 66.84 & 42.47 & 43.69 & 39.39 & 51.69 \\
& CoT & \cellcolor{green!90}75.56 & \cellcolor{green!90}59.98 & \cellcolor{green!63}68.50 & \cellcolor{green!66}58.33 & \cellcolor{green!76}63.74 & \cellcolor{green!57}41.93 & \cellcolor{green!76}61.14 \\
& S-CoT & \cellcolor{red!90}67.18 & \cellcolor{red!36}53.39 & \cellcolor{red!90}61.91 & \cellcolor{green!60}56.18 & \cellcolor{green!63}58.47 & \cellcolor{red!90}34.96 & \cellcolor{green!43}55.05 \\
& Self-Consistency & \cellcolor{green!81}75.18 & \cellcolor{green!53}56.86 & \cellcolor{green!90}69.71 & \cellcolor{green!90}67.69 & \cellcolor{green!90}69.04 & \cellcolor{green!89}44.54 & \cellcolor{green!90}63.83 \\
& Describe-Answer & \cellcolor{green!52}73.90 & \cellcolor{red!38}53.28 & \cellcolor{red!74}63.11 & \cellcolor{red!90}42.41 & \cellcolor{green!42}50.27 & \cellcolor{green!26}39.50 & \cellcolor{green!35}53.49 \\
& Self-Aggregation & \cellcolor{green!88}75.46 & \cellcolor{green!77}58.87 & \cellcolor{green!65}68.59 & \cellcolor{green!82}64.53 & \cellcolor{green!84}66.56 & \cellcolor{green!90}44.58 & \cellcolor{green!82}62.26 \\
& Self-Refinement & \cellcolor{green!65}74.49 & \cellcolor{green!86}59.62 & \cellcolor{green!63}68.50 & \cellcolor{green!72}60.64 & \cellcolor{green!75}63.14 & \cellcolor{green!36}40.25 & \cellcolor{green!74}60.87 \\
& CCoT & \cellcolor{green!90}75.56 & \cellcolor{red!90}48.00 & \cellcolor{red!71}63.38 & \cellcolor{green!34}45.83 & \cellcolor{green!42}50.13 & \cellcolor{red!71}36.26 & \cellcolor{green!30}52.65 \\
& Prompt Repetition & \cellcolor{green!70}74.71 & \cellcolor{red!73}49.70 & \cellcolor{red!32}66.29 & 42.47 & \cellcolor{green!25}43.81 & \cellcolor{red!75}35.97 & \cellcolor{red!90}50.62 \\
\midrule
\multirow{9}{*}{\rotatebox[origin=c]{90}{\textbf{Qwen3-VL-4B}}} & Baseline & 77.88 & 60.38 & 68.90 & 56.93 & 62.27 & 46.70 & 61.71 \\
& CoT & \cellcolor{red!90}74.90 & \cellcolor{green!47}61.96 & \cellcolor{green!79}74.33 & \cellcolor{green!82}75.98 & \cellcolor{green!80}79.62 & \cellcolor{green!53}48.75 & \cellcolor{green!73}68.59 \\
& S-CoT & \cellcolor{red!59}76.33 & \cellcolor{green!65}63.24 & \cellcolor{green!72}73.64 & \cellcolor{green!76}73.94 & \cellcolor{green!80}79.87 & \cellcolor{green!37}47.57 & \cellcolor{green!72}68.47 \\
& Self-Consistency & \cellcolor{red!57}76.42 & \cellcolor{green!90}65.08 & \cellcolor{green!90}75.40 & \cellcolor{green!86}77.58 & \cellcolor{green!90}82.94 & \cellcolor{green!80}50.70 & \cellcolor{green!90}71.07 \\
& Describe-Answer & \cellcolor{red!36}77.36 & \cellcolor{green!36}61.14 & \cellcolor{green!45}70.86 & \cellcolor{green!37}61.11 & \cellcolor{green!37}66.11 & \cellcolor{red!86}43.11 & \cellcolor{green!31}62.62 \\
& Self-Aggregation & \cellcolor{red!71}75.78 & \cellcolor{green!63}63.12 & \cellcolor{green!90}75.40 & \cellcolor{green!90}78.77 & \cellcolor{green!86}81.51 & \cellcolor{green!90}51.47 & \cellcolor{green!86}70.51 \\
& Self-Refinement & \cellcolor{red!61}76.23 & \cellcolor{green!41}61.52 & \cellcolor{green!79}74.33 & \cellcolor{green!80}75.49 & \cellcolor{green!82}80.49 & \cellcolor{red!48}45.35 & \cellcolor{green!70}68.19 \\
& CCoT & \cellcolor{red!54}76.55 & \cellcolor{green!78}64.23 & \cellcolor{green!31}69.49 & \cellcolor{green!35}60.42 & \cellcolor{green!36}65.66 & \cellcolor{red!90}42.90 & \cellcolor{green!33}62.81 \\
& Prompt Repetition & \cellcolor{green!90}78.35 & \cellcolor{red!90}59.57 & \cellcolor{green!37}70.09 & \cellcolor{green!33}59.46 & \cellcolor{green!29}63.62 & \cellcolor{green!37}47.58 & \cellcolor{green!27}62.03 \\
\midrule
\multirow{9}{*}{\rotatebox[origin=c]{90}{\textbf{Qwen3-VL-8B}}} & Baseline & 78.56 & 62.09 & 75.56 & 63.23 & 61.24 & 49.05 & 64.95 \\
& CoT & \cellcolor{red!41}76.14 & \cellcolor{green!28}62.47 & \cellcolor{green!26}75.65 & \cellcolor{green!81}71.87 & \cellcolor{green!90}81.03 & \cellcolor{green!41}51.48 & \cellcolor{green!56}69.77 \\
& S-CoT & \cellcolor{red!47}75.17 & \cellcolor{red!27}61.76 & \cellcolor{red!36}73.87 & \cellcolor{green!74}70.77 & \cellcolor{green!90}81.99 & \cellcolor{green!37}50.90 & \cellcolor{green!52}69.08 \\
& Self-Consistency & \cellcolor{red!36}76.81 & \cellcolor{green!42}64.69 & \cellcolor{green!38}77.59 & \cellcolor{green!90}77.47 & \cellcolor{green!90}84.87 & \cellcolor{green!73}56.49 & \cellcolor{green!77}72.99 \\
& Describe-Answer & \cellcolor{green!31}79.45 & \cellcolor{green!30}62.82 & \cellcolor{red!31}74.63 & \cellcolor{red!36}61.52 & \cellcolor{red!27}60.95 & \cellcolor{red!57}44.16 & \cellcolor{red!32}63.92 \\
& Self-Aggregation & \cellcolor{red!43}75.84 & \cellcolor{green!37}63.91 & \cellcolor{green!41}78.05 & \cellcolor{green!90}75.70 & \cellcolor{green!90}86.26 & \cellcolor{green!53}53.36 & \cellcolor{green!72}72.19 \\
& Self-Refinement & \cellcolor{red!49}74.89 & \cellcolor{red!36}60.32 & \cellcolor{red!29}74.91 & \cellcolor{green!90}74.44 & \cellcolor{green!90}82.11 & \cellcolor{red!30}48.31 & \cellcolor{green!52}69.16 \\
& CCoT & \cellcolor{red!28}78.12 & \cellcolor{green!46}65.25 & \cellcolor{red!43}72.85 & \cellcolor{green!45}66.29 & \cellcolor{red!25}61.17 & \cellcolor{green!25}49.12 & \cellcolor{green!28}65.47 \\
& Prompt Repetition & \cellcolor{green!30}79.27 & \cellcolor{green!25}62.11 & \cellcolor{green!26}75.70 & \cellcolor{green!29}63.90 & \cellcolor{red!31}60.38 & \cellcolor{red!27}48.78 & \cellcolor{green!25}65.02 \\
\end{longtable}
}

\subsection{GPT-5.2 Results}
\label{supp:gpt5.2}

To assess whether this effect also holds for strong closed-source models, we conducted experiments on MMStar and RealWorldQA using GPT-5.2, currently OpenAI’s strongest model. The results for GPT-5.2 are reported in Table \ref{tab:gpt5.2}. We find that the effect is less pronounced for GPT-5.2, but it still leads to performance degradation in some cases. For example, CoT and self-refinement reduce performance in the Coarse Perception category of MMStar, while self-refinement also degrades performance in the Fine-Grained Perception category and on RealWorldQA.

\begin{table}[h]
\centering
\caption{Results on the closed source GPT 5.2. Cell colors (\protect\colorbox{red!25}{degradation} or \protect\colorbox{green!25}{improvement}) indicate change relative to the baseline on each dataset.}
\label{tab:gpt5.2}
\begin{tabular}{lrrrrrrrr}
\toprule
\textbf{Method} &
\rotatebox{90}{\makecell{Coarse \\ Perception}} &
\rotatebox{90}{\makecell{Fine-grained \\ Perception}} &
\rotatebox{90}{\makecell{Instance \\ Reasoning}} &
\rotatebox{90}{\makecell{Logical \\ Reasoning}} &
\rotatebox{90}{\makecell{Math}} &
\rotatebox{90}{\makecell{Science}} &
\rotatebox{90}{\makecell{Overall}} &
\rotatebox{90}{\makecell{RealWorldQA}} \\
\midrule
Baseline
& 76.29 & 58.54 & 73.85 & 61.55 & 52.21 & 52.53 & 62.49 & 73.59 \\

CoT
& \cellcolor{red!26}73.50
& \cellcolor{green!26}61.38
& \cellcolor{green!22}74.81
& \cellcolor{green!43}72.90
& \cellcolor{green!70}82.38
& \cellcolor{green!39}62.06
& \cellcolor{green!37}71.17
& 73.59 \\

S-CoT
& \cellcolor{green!25}78.87
& \cellcolor{green!29}63.03
& \cellcolor{green!25}76.48
& \cellcolor{green!49}75.82
& \cellcolor{green!70}77.58
& \cellcolor{green!44}64.30
& \cellcolor{green!40}72.68
& \cellcolor{green!28}77.78 \\

Self-Refinement
& \cellcolor{red!33}69.70
& \cellcolor{red!21}58.10
& \cellcolor{green!29}78.47
& \cellcolor{green!44}73.67
& \cellcolor{green!70}82.51
& \cellcolor{green!46}65.43
& \cellcolor{green!38}71.31
& \cellcolor{red!23}71.90 \\

Self-Aggregation
& \cellcolor{green!23}77.74
& \cellcolor{green!21}59.05
& \cellcolor{green!31}79.31
& \cellcolor{green!55}78.91
& \cellcolor{green!70}79.07
& \cellcolor{green!41}63.17
& \cellcolor{green!41}72.87
& \cellcolor{green!26}76.60 \\

Prompt Repetition 
& \cellcolor{green!21}76.85
& \cellcolor{red!28}54.59
& \cellcolor{red!23}72.52
& \cellcolor{green!26}64.48
& \cellcolor{green!28}55.97
& \cellcolor{red!26}49.75
& \cellcolor{red!20}62.36
& \cellcolor{green!23}75.03 \\

\bottomrule
\end{tabular}
\end{table}

\section{Token Cutoff}
\label{supp:token_cutoff}

\noindent We report the score difference ($\Delta$) for the perception benchmarks when the reasoning chain is truncated at 300 tokens. Specifically, $\Delta = \text{score}{300} - \text{score}{1024}$. The results are shown in Table \ref{supptab:token_budget_aokvqa} for the A-OKVQA benchmark and in Table \ref{supptab:token_budget_mmstar_cat} for the perception categories of MMStar. We also report token cutoff results for RealWorldQA and HallusionBench for InternVL-3.5-8B and 38B models in Table \ref{supptab:token_budget_intenrvl}. We report $\Delta = \text{score}_{300} - \text{score}_{1024}$. \colorbox{green!30}{\strut improvement} indicates a positive $\Delta$ (300-token early stopping improves performance). \colorbox{red!30}{\strut degradation} indicates negative $\Delta$. As shown in the Tables, most methods perform better under the 300-token budget than under the 1024-token budget, or at least do not cause degradation. This further supports the idea that short answers are more effective for perceptual tasks, whereas overly verbose responses can cause the LVLM to lose focus, hallucinate, and ultimately produce incorrect answers (Takeaway 2).

\begin{table*}[h]
\centering
\caption{Early stopping at token 300 on A-OKVQA}
\setlength{\tabcolsep}{4pt}
\renewcommand{\arraystretch}{1.08}
\small
\label{tab:delta_300_1024_aokvqa_transposed}

\scalebox{0.85}{%
\begin{tabular}{l c c c c c c c c}
\toprule
\textbf{Model}
& \rotatebox{90}{\textbf{CoT}}
& \rotatebox{90}{\textbf{CCoT}}
& \rotatebox{90}{\textbf{S-CoT}}
& \rotatebox{90}{\textbf{Prompt Rep.}}
& \rotatebox{90}{\textbf{Desc.-Ans.}}
& \rotatebox{90}{\textbf{Self-Cons.}}
& \rotatebox{90}{\textbf{Self-Refine.}}
& \rotatebox{90}{\textbf{Self-Agg.}} \\
\midrule

Qwen3-VL-4B
& \cellcolor{red!15} -0.35
& \cellcolor{green!15} 0.17
& \cellcolor{red!15} -0.61
& 0.00
& \cellcolor{red!15} -0.35
& \cellcolor{red!15} -0.70
& \cellcolor{red!15} -0.35
& \cellcolor{red!15} -0.88 \\

Qwen3-VL-8B
& \cellcolor{red!15} -0.52
& \cellcolor{green!30} 0.61
& \cellcolor{green!15} 0.17
& 0.00
& \cellcolor{green!30} 0.53
& \cellcolor{green!15} 0.08
& \cellcolor{red!15} -0.09
& \cellcolor{green!15} 0.35 \\

Qwen3-VL-32B
& \cellcolor{red!15} -0.18
& \cellcolor{green!15} 0.43
& \cellcolor{green!30} 0.52
& 0.00
& \cellcolor{red!15} -0.17
& \cellcolor{red!15} -0.18
& \cellcolor{green!30} 0.79
& \cellcolor{red!15} -0.26 \\

Qwen2.5-VL-3B
& 0.00
& 0.00
& \cellcolor{green!15} 0.35
& 0.00
& \cellcolor{green!15} 0.09
& \cellcolor{red!15} -0.70
& 0.00
& \cellcolor{red!15} -0.96 \\

Qwen2.5-VL-7B
& \cellcolor{green!15} 0.09
& \cellcolor{green!15} 0.26
& 0.00
& 0.00
& \cellcolor{green!15} 0.18
& \cellcolor{green!30} 0.52
& \cellcolor{red!15} -0.09
& \cellcolor{red!15} -0.52 \\

Qwen2.5-VL-32B
& 0.00
& \cellcolor{green!15} 0.09
& \cellcolor{red!15} -0.17
& 0.00
& \cellcolor{red!15} -0.70
& \cellcolor{green!30} 0.52
& \cellcolor{green!30} 0.53
& \cellcolor{green!15} 0.18 \\

\midrule

Molmo2-4B
& 0.00
& 0.00
& 0.00
& 0.00
& \cellcolor{red!15} -0.26
& \cellcolor{red!15} -0.61
& 0.00
& \cellcolor{red!15} -0.88 \\

Molmo2-8B
& 0.00
& \cellcolor{green!15} 0.08
& 0.00
& 0.00
& 0.00
& \cellcolor{red!15} -0.18
& 0.00
& \cellcolor{green!15} 0.44 \\

\bottomrule
\end{tabular}%
}
\label{supptab:token_budget_aokvqa}
\end{table*}

\begin{table*}[h]
\centering
\caption{Early stopping at token 300 on the perceptual categories of MMStar}
\setlength{\tabcolsep}{3.2pt}
\renewcommand{\arraystretch}{1.08}
\small

\scalebox{0.99}{%
\begin{tabular}{l l c c c}
\toprule
\textbf{Model} & \textbf{Method} & \makecell{\textbf{Coarse} \\ \textbf{Perception}} & \makecell{\textbf{Fine-grained} \\ \textbf{Perception}} & \makecell{\textbf{Science \&} \\ \textbf{Technology}} \\
\midrule

\multirow{4}{*}{Qwen3-VL-4B}
& CoT       & \cellcolor{red!15} -0.60 & \cellcolor{green!15} 0.37 & \cellcolor{white} 0.00 \\
& S-CoT       & \cellcolor{green!30} 1.08 & \cellcolor{green!30} 0.92 & \cellcolor{red!60} -3.47 \\
& Self-Ref. & \cellcolor{red!30} -1.28 & \cellcolor{red!15} -0.73 & \cellcolor{red!15} -0.04 \\
& Self-Agg. & \cellcolor{green!30} 1.63 & \cellcolor{red!15} -0.53 & \cellcolor{red!60} -4.35 \\
\midrule

\multirow{4}{*}{Qwen3-VL-8B}
& CoT       & \cellcolor{red!45} -2.68 & \cellcolor{green!15} 0.09 & \cellcolor{green!30} 1.72 \\
& S-CoT       & \cellcolor{green!60} 3.01 & \cellcolor{red!30} -1.01 & \cellcolor{green!45} 2.96 \\
& Self-Ref. & \cellcolor{red!45} -2.79 & \cellcolor{green!30} 1.68 & \cellcolor{green!30} 1.81 \\
& Self-Agg. & \cellcolor{green!30} 1.46 & \cellcolor{red!30} -1.35 & \cellcolor{red!30} -1.53 \\
\midrule

\multirow{4}{*}{Qwen3-VL-32B}
& CoT       & \cellcolor{red!45} -2.17 & \cellcolor{green!30} 1.00 & \cellcolor{green!60} 4.85 \\
& S-CoT       & \cellcolor{green!30} 1.51 & \cellcolor{green!30} 1.41 & \cellcolor{red!15} -0.22 \\
& Self-Ref. & \cellcolor{green!30} 1.69 & \cellcolor{green!30} 1.45 & \cellcolor{green!60} 4.89 \\
& Self-Agg. & \cellcolor{green!15} 0.84 & \cellcolor{red!15} -0.07 & \cellcolor{green!45} 2.10 \\
\midrule

\multirow{4}{*}{Qwen2.5-VL-7B}
& CoT       & \cellcolor{red!30} -1.93 & \cellcolor{green!30} 1.51 & \cellcolor{red!45} -2.48 \\
& S-CoT       & \cellcolor{green!30} 1.42 & \cellcolor{red!45} -2.06 & \cellcolor{red!45} -2.56 \\
& Self-Ref. & \cellcolor{red!45} -2.39 & \cellcolor{red!15} -0.89 & \cellcolor{red!30} -1.40 \\
& Self-Agg. & \cellcolor{green!15} 0.32 & \cellcolor{green!60} 3.45 & \cellcolor{green!30} 1.27 \\
\midrule

\multirow{4}{*}{Qwen2.5-VL-32B}
& CoT       & \cellcolor{red!30} -1.12 & \cellcolor{red!15} -0.28 & \cellcolor{green!45} 2.03 \\
& S-CoT       & \cellcolor{red!45} -2.86 & \cellcolor{red!60} -4.33 & \cellcolor{green!15} 0.25 \\
& Self-Ref. & \cellcolor{green!60} 4.90 & \cellcolor{red!15} -0.09 & \cellcolor{green!75} 7.28 \\
& Self-Agg. & \cellcolor{green!30} 1.20 & \cellcolor{green!45} 2.03 & \cellcolor{green!60} 3.30 \\
\midrule

\multirow{4}{*}{Qwen2.5-VL-72B}
& CoT       & 0.00 & 0.00 & \cellcolor{green!15} 0.35 \\
& S-CoT       & 0.00 & 0.00 & \cellcolor{red!30} -1.53 \\
& Self-Ref. & \cellcolor{green!30} 1.28 & \cellcolor{red!15} -0.28 & \cellcolor{green!60} 4.92 \\
& Self-Agg. & \cellcolor{red!45} -2.40 & \cellcolor{red!45} -2.10 & \cellcolor{red!45} -2.17 \\
\bottomrule
\end{tabular}%
}
\footnotesize
\label{supptab:token_budget_mmstar_cat}
\end{table*}

\begin{table}
\centering
\caption{Early stopping at token 300 on RealWorldQA and HallusionBench on InternVL-3.5. }
\setlength{\tabcolsep}{4pt}
\renewcommand{\arraystretch}{1.08}
\small
\label{tab:delta_300_1024_realworldqa_hallusionbench}

\scalebox{0.99}{%
\begin{tabular}{l l c c c}
\toprule
\textbf{Benchmark} & \textbf{Model}
& \rotatebox{90}{\textbf{CoT}}
& \rotatebox{90}{\textbf{S-CoT}}
& \rotatebox{90}{\textbf{Self-Agg.}} \\
\midrule

\multirow{2}{*}{RealWorldQA}
& InternVL-3.5-8B
& 0.00
& 0.00
& \cellcolor{green!30} 1.70 \\

& InternVL-3.5-38B
& 0.00
& \cellcolor{red!15} -0.13
& \cellcolor{green!15} 0.39 \\

\midrule

\multirow{2}{*}{HallusionBench}
& InternVL-3.5-8B
& \cellcolor{red!15} -0.52
& \cellcolor{red!15} -0.42
& \cellcolor{red!15} -0.22 \\

& InternVL-3.5-38B
& 0.00
& \cellcolor{red!15} -0.63
& \cellcolor{green!45} 2.84 \\

\bottomrule
\end{tabular}%
}
\label{supptab:token_budget_intenrvl}
\end{table}

\section{Additional Attention Dynamics Analysis}
\label{supp:img_attn}
\begin{figure}
    \centering
    \includegraphics[width=0.8\linewidth]{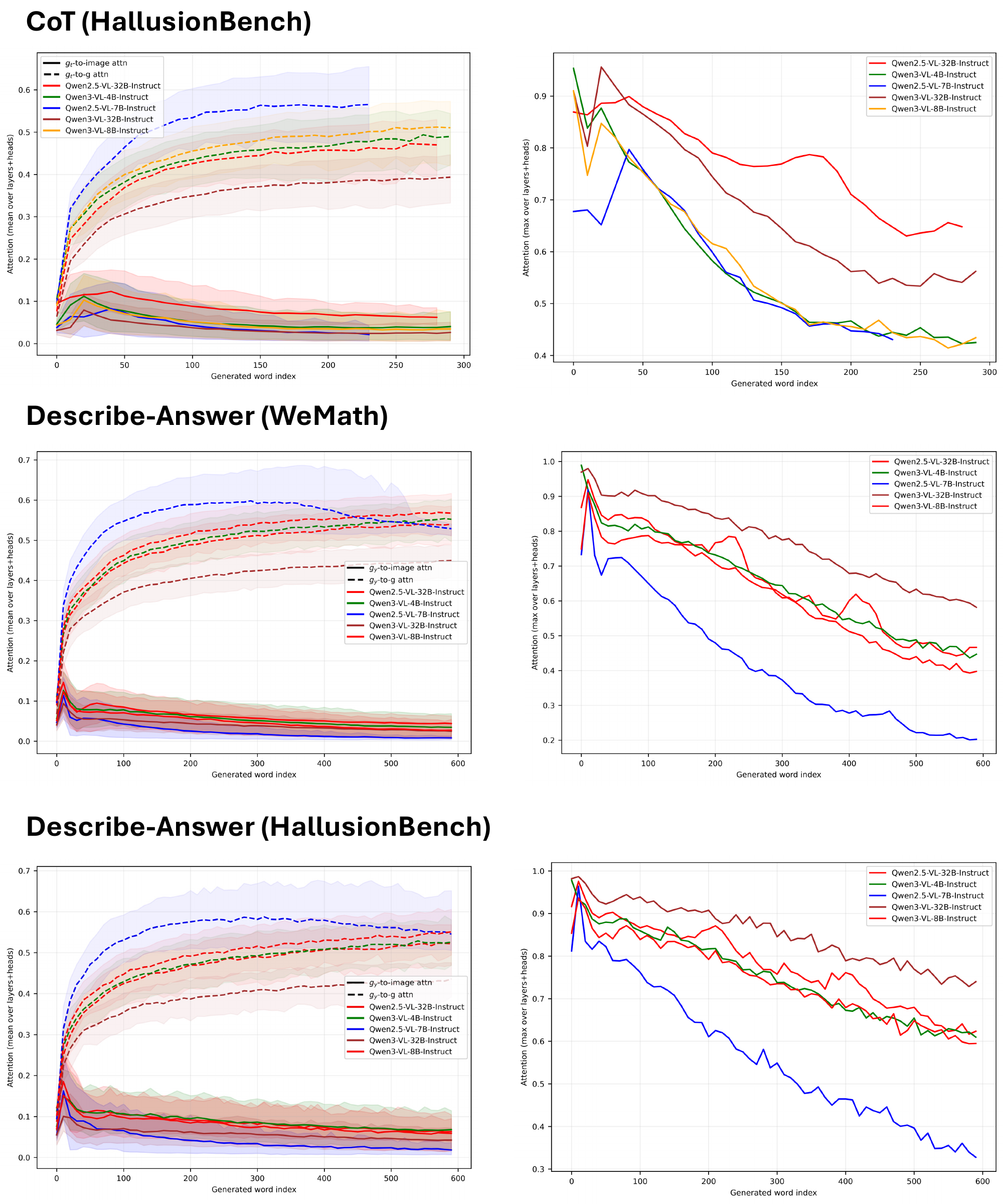}
    \caption{Attention Dynamics during Reasoning for additional datasets and methods. For each generated token $g_y$ , we measure attention to image tokens (solid curve) and previously generated tokens (dashed curve). (Left) Average attention across layers, heads, and samples for WeMath. (Right) Upper bound on image attention, defined as the maximum attention to image tokens across layers and heads}
    \label{fig:additional_attention_analysis}
\end{figure}

In the main manuscript, we presented attention dynamics during reasoning for chain-of-thought prompting on WeMath. In Figure \ref{fig:additional_attention_analysis} (Top), we complement this with  chain-of-thought prompting on HallusionBench. We also show the attention dynamics for both WeMath and HallusionBench for the Describe-Answer method. All dynamics share the same trend as those in the main manuscript, verifying that our claim generalizes to other methods and datasets.

\section{KV Cache Drop (Additional Datasets)}
\label{supp:kv_cache_drop_additional}
\begin{figure}
    \centering
    \includegraphics[width=0.8\linewidth]{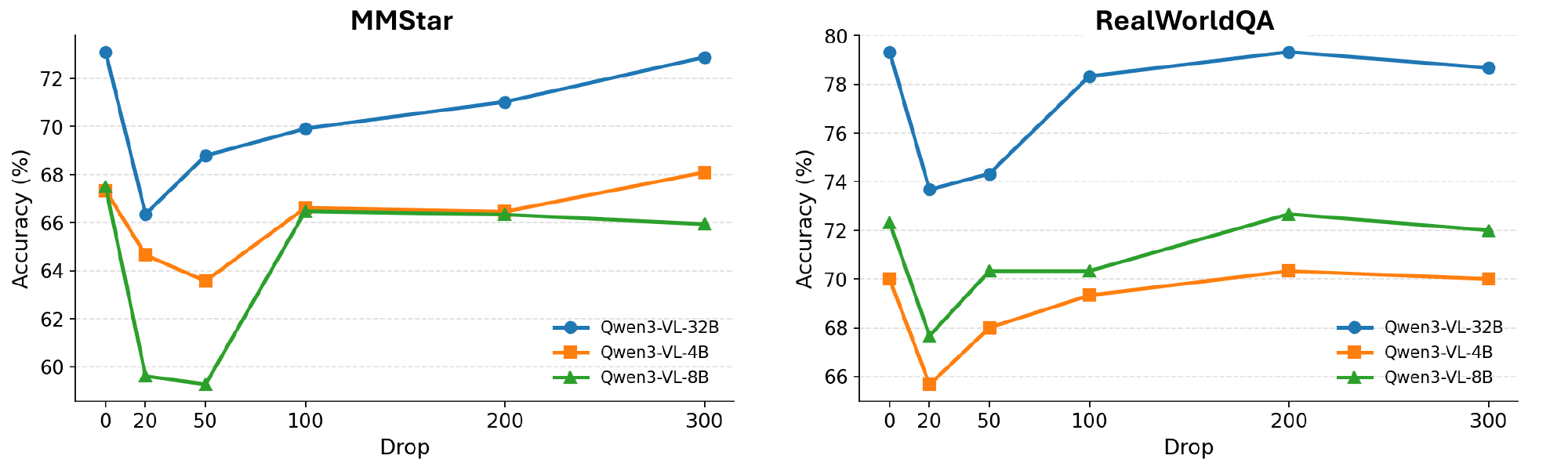}
    \caption{Image KV Cache Dropping for MMStar and RealWorldQA}
    \label{fig:kv_cache_drop_supp}
\end{figure}

We present the image token dropping causal intervention results for MMStar and RealWorldQA benchmarks in Figure \ref{fig:kv_cache_drop_supp}. 

\section{KV Cache Drop FLOPs Savings}
\label{supp:kv_cache_drop_flops}
\begin{figure}
    \centering
    \includegraphics[width=0.8\linewidth]{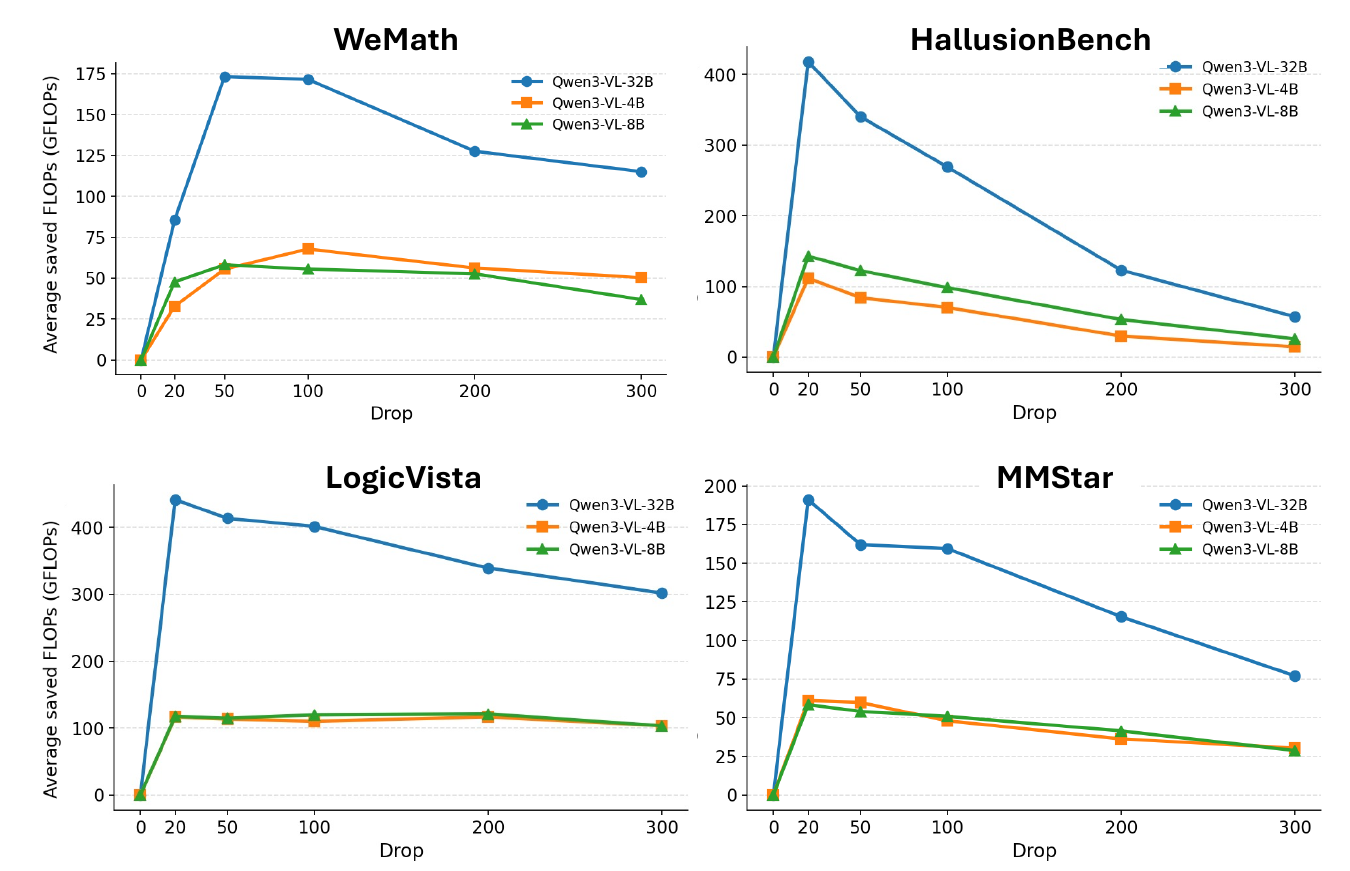}
    \caption{GFLOPs saved (y-axis) when performing Image KV Cahce Dropping at different drop steps (x-axis).}
    \label{fig:dop_kv_cache_flops}
\end{figure}

We complement our analysis on image token dropping with an analysis on computational savings. In Figure \ref{fig:dop_kv_cache_flops}, we present savings in GFLOPs (y-axis) at different generation step positions (x-axis). The FLOPs are calculated theoretically for attention blocks across all layers, since other modules such as MLP have no effect as they do not process stored KV cache tokens (including the image tokens that are dropped). As shown, early dropping leads to the highest savings in computation, but at the expense of performance degradation due to dropping image tokens during the front-loading phase. From the main manuscript, dropping image tokens after 200 tokens have been generated leads to almost no degradation in performance. This translates to approximately 125 GFLOPs saved for WeMath, HallusionBench and MMStar, and around 325 GFLOPs saved for LogicVista. 

\clearpage
\section{Think prompts}
\label{supp:thinking_prompts}

We report the \texttt{think-prompt} for each method below. 
\newline

\noindent \textbf{CoT:} 
\begin{verbatim}
Think step by step.
\end{verbatim}

\noindent \textbf{S-CoT:} 
\begin{verbatim}
1. Understand the question and break it down into independent 
concepts and components. 
2. Then outline relevant information for each. 
3. Apply logical reasoning to derive conclusions from the information 
and provide a step-by-step articulation of your reasoning process. 
4. Summarize the main points that are relevant to answering the question. 
\end{verbatim}

\noindent \textbf{PaS:} 
\begin{verbatim}
First understand the question and devise a plan to solve the question. 
Then, carry out the plan and solve the question step by step.
\end{verbatim}

\noindent \textbf{Self-Aggregation:} 
\begin{verbatim}
Question: {question}
You will read multiple solutions (may be redundant or wrong):
{candidates}
Using the image, review the solutions and analyze their consistency 
and correctness. Then combine the useful and correct ideas, 
think and reason about them, and produce a final correct solution 
to the question with step-by-step reasoning.
At the end, provide your final answer inside <answer>...</answer> tags.
\end{verbatim}

\noindent \textbf{Self-Refinement:} 
\begin{verbatim}
Question: {question}
You previously wrote this solution to answer the question 
(which may contain mistakes):{prev_chain}
Using the image and the provided solution, your task is to improve 
the solution you wrote. 
Identify errors, reuse any correct information, and write a better 
solution to the question.
At the end, provide your final answer inside <answer>...</answer> tags. 
Refined Solution:
\end{verbatim}

\textbf{Describe-Answer:} 
\begin{verbatim}
Describe this image in full detail. Describe all the objects 
and attributes and their relationship with each other.
\end{verbatim}

\textbf{CCoT:} 
\begin{verbatim}
For the provided image and its associated question, generate a scene 
graph in JSON format that includes the following:
1. Objects that are relevant to answering the question
2. Object attributes that are relevant to answering the question
3. Object relationships that are relevant to answering the question

Scene Graph:
\end{verbatim}

\noindent Note that for prompt repetition, the \texttt{think-prompt} is just the \texttt{question}. Moreover, Self-Consistency does not have a specific \texttt{think-prompt} as it is just involves sampling multiple responses for the CoT method, which means it uses the \texttt{think-prompt} of the CoT method for each sample. 

\section{Post Prompts}
\label{supp:post_prompts}
We report the \texttt{post-prompt} used for each benchmark. Table \ref{tab:benchmark_overview} summarizes the task(s) for each of the benchmarks.
\newline
\newline
\noindent For MCQ tasks, we use the following post prompts:
\newline
\newline
\noindent \textbf{First Round:} \texttt{Answer the question by providing the correct option's letter from the given choices inside <answer>...</answer> tags."}

\noindent If the first round fails to include the answer extraction tags (<answer></answer>), or the method requires a second round (Describe-Answer, CCoT), we use the following second round post-prompt:

\noindent \textbf{Second Round:} \texttt{Answer only with the option's letter from the given choices directly.}
\newline
\newline
\noindent For Open-Form answer tasks, we use the following post prompt:
\newline
\newline
\noindent \textbf{First Round:} \texttt{Answer the question using a single word, number or short phrase inside <answer>...</answer> tags}

\noindent If the first round fails to include the answer extraction tags (<answer></answer>), or the method requires a second round, we use the following second round post-prompt:

\noindent \textbf{Second Round:} \texttt{Answer the question using a single word, number or short phrase.}

\begin{table}[h]
\centering
\caption{Benchmark Overview}
\label{tab:benchmark_overview}
\renewcommand{\arraystretch}{1.2}
\begin{tabular}{ll}
\hline
\textbf{Benchmark} & \textbf{Task} \\
\hline
MMStar & MCQ \\
RealWorldQA & MCQ, Open-Ended \\
HallusionBench & Yes/No \\
WeMath & MCQ \\
LogicVista & MCQ \\
\hline
\end{tabular}
\end{table}

\section{Human validation of the LLM Judge}
\label{supp:human_validation}

To assess the reliability of the LLM judge, we conduct a human evaluation on 250 randomly sampled, non-overlapping instances drawn from all five benchmarks. Five annotators each evaluated 50 samples. For each instance, annotators were given the question $q$ and the generated rationale $r$ (without access to the image, external information, nor the ground-truth answer $x$) and were asked to predict the final answer based solely on the text rationale $r$, mirroring the judge protocol. Table~\ref{tab:judge_validation} reports pairwise agreement rates between human annotations, LVLM predictions, the LLM judge, and ground truth. The LLM judge achieves accuracy comparable to the LVLM when evaluated against ground truth (76.4\% vs.\ 75.9\%) and shows strong agreement with both the LVLM (94.3\%) and human annotators (83.3\%). These results indicate that the judge’s decisions are highly consistent with both model outputs and human assessments, supporting its use as a reliable proxy for evaluating rationale sufficiency.

\begin{table}[h]
  \centering
  \caption{Pairwise agreement rates between human annotators, LVLM predictions, the LLM judge and ground truth (GT).}
  \label{tab:judge_validation}
  \small
  \begin{tabular}{l c}
    \toprule
    \textbf{Comparison} & \textbf{Agreement (\%)} \\
    \midrule
    Human -- GT & 69.5 \\
    VLM -- GT & 75.9 \\
    LLM Judge -- GT & 76.4 \\
    Human -- VLM & 86.2 \\
    Human -- LLM Judge & 83.3 \\
    VLM -- LLM Judge & \textbf{94.3} \\
    \bottomrule
  \end{tabular}
\end{table}

\section{Method Parameters}
\label{supp:method_params}
We use greedy decoding (temperature = 0) for all methods, except for those that require diverse sampling, which are listed below:

\noindent \textbf{Self-Consistency:} we used a temperature of 0.8, with nucleus sampling (\textit{top-p} = 0.95). We use 10 steps.

\noindent \textbf{Self-Aggregation:} we used a temperature of 0.8, with nucleus sampling (\textit{top-p} = 0.95). We use 4 candidates.

\section{Other model failures}
\label{supp:model_failures}
\begin{figure}
    \centering
    \includegraphics[width=1\linewidth]{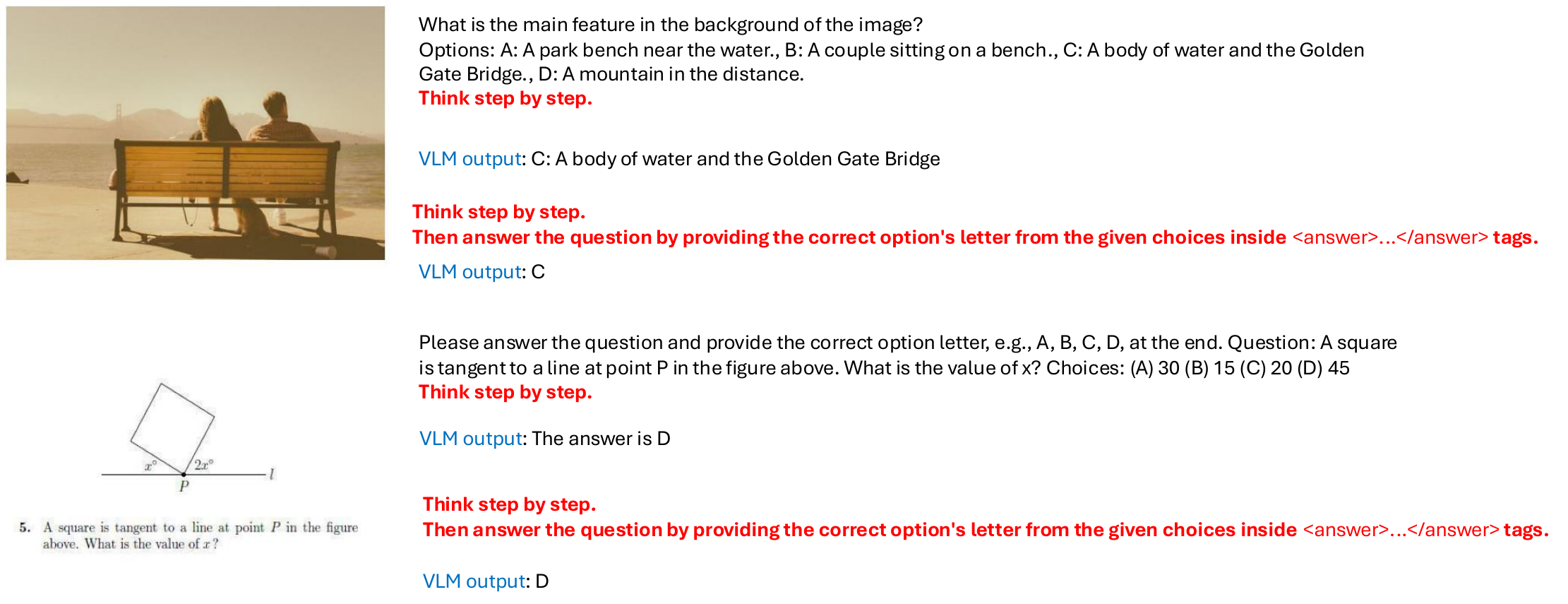}
    \caption{Failure of LLaVA-OneVision-7B to follow instructions, behaving more like a specialized VQA system than a conversational agent.}
    \label{fig:failure_llava}
\end{figure}

\begin{figure}
    \centering
    \includegraphics[width=1\linewidth]{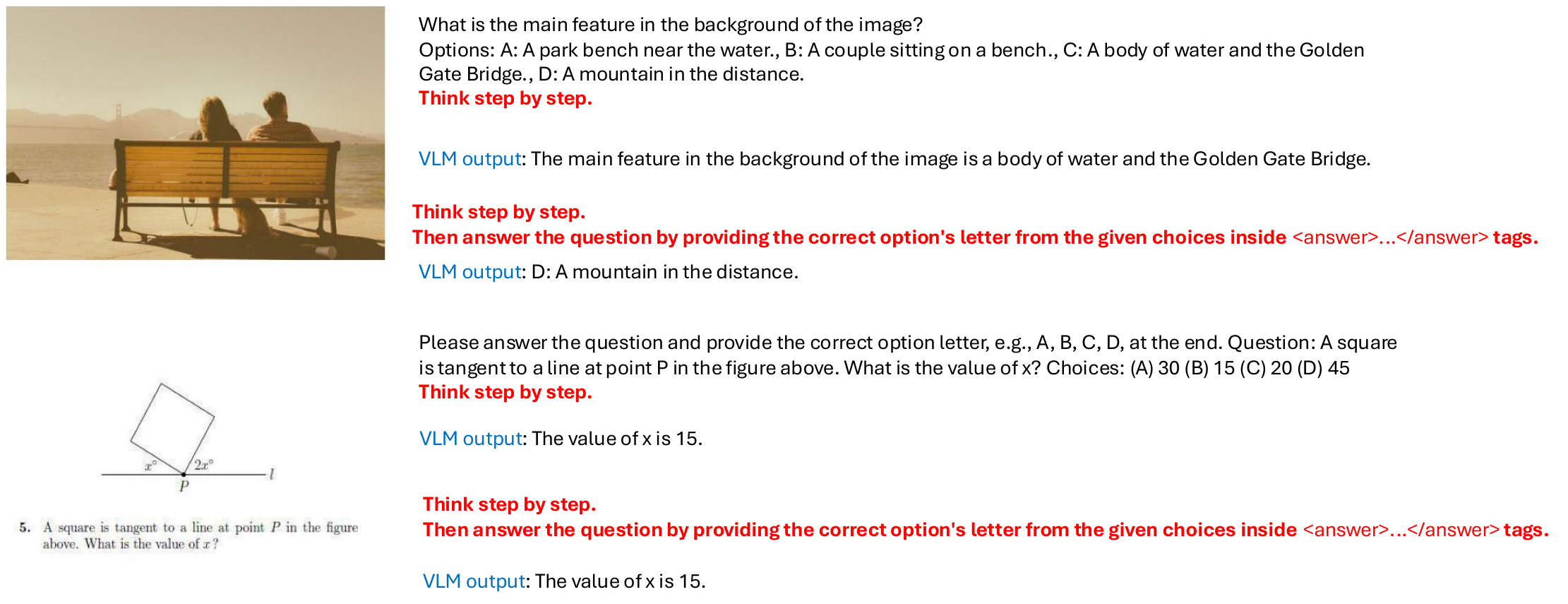}
    \caption{Failure of SmolVLM2-2.2B to follow instructions, behaving more like a specialized VQA system than a conversational agent.}
    \label{fig:failure_smolvlm}
\end{figure}

We also examined the behavior of other LVLMs, specifically LLaVA-OneVision-7B and SmolVLM2-2.2B. We found that these models often exhibit poor instruction-following behavior: they answer the questions directly while ignoring the \texttt{think-prompt}. They behave more like specialized VQA systems than instruction-following conversational agents. We hypothesized that the failure could be because the \texttt{post-prompt} instruction may distract the LVLM from the \texttt{think-prompt} instruction, so we tested a case where we only provide the \texttt{think-prompt} instruction (without the \texttt{post-prompt} instruction), and another case where we provide both. We used the simplest CoT method to test this. Figure \ref{fig:failure_llava} illustrates the behavior for LLaVA-OneVision-7B, while Figure \ref{fig:failure_smolvlm} shows the behavior for SmolVLM2-2.2B. In both cases, both models fail. 

\section{Attention Curves Plot Details}
\label{supp:attn_plot_details}
We provide details for the plot used in Figure 4 in the main manuscript. Because different samples produce variable-length reasoning traces, we align them by token index via padding and compute aggregate statistics while ignoring padded positions so only valid tokens contribute. We additionally report results only where a minimum number of samples still contribute at that token position, preventing the far tail from being driven by a handful of unusually long generations. As a result, curves can terminate at different token indices when sample support drops below this threshold (e.g., due to shorter generations or capped decoding). We considered the threshold to be be at least 20\% of samples at a token index, and with at least 5 samples. 

\clearpage

\clearpage
\section{LLM-as-a-Judge Prompts}
\label{supp:llm-judge-prompts}
We provide the prompts used for LLM-as-a-Judge experiments for both the Rational Sufficiency and Rational Dynamics.

Rational Sufficiency: 

\begin{verbatim}
You are an evaluator for a vision language model. 
Your job is to judge whether the given rationale contains enough 
information to answer the question WITHOUT seeing the image.

You will receive:
- question (Q)
- rationale (R)

IMPORTANT:
- Do NOT assume any visual information not explicitly stated in R.
- Do NOT use outside world knowledge.
- Do NOT assume any visual information only judge based on the text 
provided.
- Answer the question using ONLY R.
- Your "predicted_answer_from_rationale" MUST be plain text only:
  - NO XML/HTML tags (e.g., no <answer>...</answer>)
  - NO markdown formatting
  - NO surrounding quotes beyond normal JSON string quoting

Scoring:
- Provide a single integer score from 0 to 10:
  0 = impossible to answer from R (missing key info / pure fluff)
  10 = fully answerable from R, with clear support

Return ONLY valid JSON with exactly these keys and nothing else:
{
  "score": <int 0..10>,
  "predicted_answer_from_rationale": "<plain text answer>",
  "explanation": "<brief explanation of how you arrived at the score 
  and predicted answer>"
}
\end{verbatim}

Rational Dynamics:

\begin{verbatim}
You are an evaluator for a vision language model. 
Your task is to analyze the rationale (R) at the sentence level.
 
You will receive:
- question (Q)
- rationale (R)
- ground-truth answer (GT)
 
IMPORTANT RULES:
- Split R into sentences in the order they appear.
- Keep only sentences that contain an actual claim, observation, 
comparison, or inference.
- Remove any structural markers or headers, including lines like:  
"Step 1:", "Conclusion:", "Main points:", 
"Logical reasoning:", "1.", "2.", "-", "*", "•", "<answer>"
- Do NOT judge factual correctness.
- Evaluate sentences sequentially.
- Do NOT use outside world knowledge.
- Evaluate only per specific sentence and relative to GT.
 
For each sentence i:
 
answer_signal:
   Based only on reasoning from this sentence,
   what answer does the sentence currently support?
   - "yes"
   - "no"
   - "uncertain"
 
Return ONLY valid JSON with this structure:
 
{
  "sentences": [
    {
      "index": 1,
      "sentence_text": "the actual text of the sentence",
      "answer_signal": "yes" | "no" | "uncertain"
    },
    ...
  ]
}

\end{verbatim}

\end{document}